\documentclass[11pt, a4paper, logo, onecolumn, copyright]{googledeepmind} %

\usepackage{amsmath,amsfonts,bm}

\def\eqref#1{equation~\ref{#1}}

\def\1{\bm{1}}

\DeclareMathAlphabet{\mathsfit}{\encodingdefault}{\sfdefault}{m}{sl}
\SetMathAlphabet{\mathsfit}{bold}{\encodingdefault}{\sfdefault}{bx}{n}

\usepackage{hyperref}
\hypersetup{
    colorlinks=true,
    linkcolor=orange,
    filecolor=magenta,      
    urlcolor=orange,
    citecolor=orange,
}
\usepackage{enumitem}
\usepackage[capitalise, nameinlink]{cleveref}
\usepackage{url}
\usepackage{graphicx}
\usepackage{subcaption}
\usepackage{xspace}
\usepackage{booktabs}       %
\usepackage{amsfonts}       %
\usepackage{nicefrac}       %
\usepackage{microtype}      %
\usepackage{amsmath,amsthm}
\usepackage{wrapfig}
\usepackage[font=small]{caption} 
\usepackage{listings}
\usepackage{algorithm}
\usepackage[noend]{algorithmic}
\usepackage{bbm}
\usepackage{mdframed}
\usepackage{pdfx}
\usepackage{natbib}
\setcitestyle{square,numbers,comma}
\usepackage{abstract}  %

\newcommand{\vlm}{\textrm{VLM}}

\title{Vision Language Models are In-Context Value Learners}

\author{Yecheng Jason Ma$^{\dagger,1,2}$, Joey Hejna$^{1,3}$, Ayzaan Wahid$^{1}$, Chuyuan Fu$^{1}$, Dhruv Shah$^{1}$, Jacky Liang$^{1}$, Zhuo Xu$^{1}$, Sean Kirmani$^{1}$, Peng Xu$^{1}$, Danny Driess$^{1}$, Ted Xiao$^{1}$, Jonathan Tompson$^{1}$, Osbert Bastani$^{2}$, Dinesh Jayaraman$^{2}$, Wenhao Yu$^{1}$, Tingnan Zhang$^{1}$, Dorsa Sadigh$^{1,3}$, Fei Xia$^{1}$ \\
$^1$Google DeepMind, $^2$University of Pennsylvania, $^3$Stanford University\\
Correspond to: \texttt{jasonyma@seas.upenn.edu, xiafei@google.com}\\
\normalsize{Website and Interactive Demo: \href{http://generative-value-learning.github.io}{generative-value-learning.github.io}}
}

\newcommand{\ourmethod}{GVL\xspace}
\begin{document}

\maketitle
\vspace{-0.5in}

\begin{abstract}
Predicting temporal progress from visual trajectories is important for intelligent robots that can learn, adapt, and improve. However, learning such progress estimator, or temporal value function, across different tasks and domains requires both a large amount of diverse data and methods which can scale and generalize. To address these challenges, we present Generative Value Learning (\ourmethod), a universal value function estimator that leverages the world knowledge embedded in vision-language models (VLMs) to predict task progress. Naively asking a VLM to predict values for a video sequence performs poorly due to the strong temporal correlation between successive frames. Instead, \ourmethod poses value estimation as a temporal ordering problem over shuffled video frames; this seemingly more challenging task encourages VLMs to more fully exploit their underlying semantic and temporal grounding capabilities to differentiate frames based on their perceived task progress, consequently producing significantly better value predictions. Without any robot or task specific training, \ourmethod can in-context zero-shot and few-shot predict effective values for more than 300 distinct real-world tasks across diverse robot platforms, including challenging bimanual manipulation tasks. Furthermore, we demonstrate that \ourmethod permits flexible multi-modal in-context learning via examples from heterogeneous tasks and embodiments, such as human videos. The generality of \ourmethod enables various downstream applications pertinent to visuomotor policy learning, including dataset filtering, success detection, and advantage-weighted regression -- all without any model training or finetuning.
\end{abstract}

\section{Introduction}

Predicting temporal progress from visual trajectories is an important task for embodied agents that interact with the physical world. A robot capable of generalizable progress estimation can in principle discern desirable and undesirable behaviors to learn visuomotor skills in new environments. This is most often studied in reinforcement learning literature~\citep{schaul2015universal}, where progress estimation is equivalent to universal value learning under specific choices of reward function. However, universal value estimation comes with a number of key challenges: (1) broad \emph{generalization} to new tasks and scenes, (2) the ability to \emph{accurately estimate state} in partially observed environments, and (3) temporal \emph{consistency} (i.e. satisfying the Bellman equation) over long horizons. Most existing methods trained on relatively small amounts of vision-only data \citep{chen2021learning,ma2022vip,ahn2022can} lack the semantic, spatial, and temporal understanding needed to ground task progress in the space-time manifold of video, preventing generalization. Moreover, they often reason over single frames, inducing a high-degree of uncertainty in partially observed environments which in turn can effect the consistency of predictions for poorly estimated states. However, these challenges are not insurmountable: modern vision language models (VLMs) exhibit marked generalization and reasoning capabilities, potentially making them useful for value estimation.

\begin{figure}[htb]
\centering 
\includegraphics[width=0.9\linewidth]{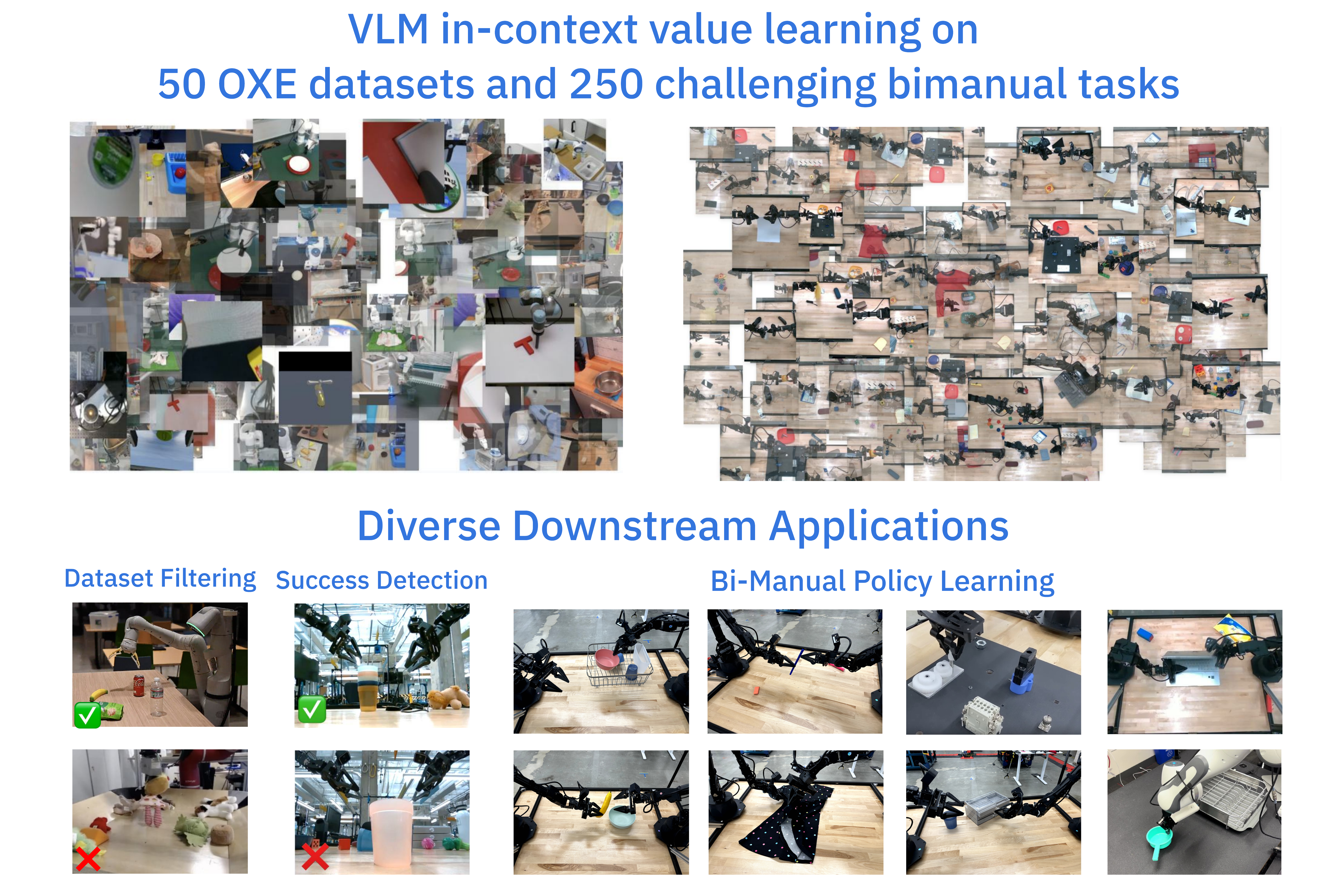}
\caption{\textbf{Result highlights.} \ourmethod can effectively zero-shot and few-shot predict task progress on diverse and challenging real-world tasks; these capabilities enable expansive set of downstream applications, including dataset filtering, success detection, and policy learning.}
\label{fig:results}
\end{figure}

Though not often considered as candidates for value estimation, VLMs excel at its aforementioned core challenges. First, state-of-the-art VLMs have exhibited strong spatial reasoning and temporal understanding capabilities across various vision tasks~\citep{nag2022zero,chen2024spatialvlm, hong20233d, gao2024physically}, allowing them to \emph{generalize} to novel scenarios. Second, large transformer-based VLMs have the requisite context window \citep{reid2024gemini} to reason over large amounts of historical information to \emph{accurately estimate state} from observation sequences when predicting task progress. Finally, VLMs make predictions \textit{auto-regressively}, meaning they commit to their own outputs as inputs for subsequent predictions, imposing \emph{consistency} constraints on long generations. For example, a VLM is unlikely to estimate that a task is 50\% completed if it already has a 50\% completion prediction in context. However, how exactly a VLM should be used to predict values is unclear. Empirically, we find that simply placing a video in-context and prompting the model to return progress predictions for each frame fails -- our analysis suggests strong temporal correlations between successive frames often cause VLMs to produce uninformative monotonic values that disregard the actual quality of the trajectory and differences between frames (\cref{sec:experiments}) -- and a different approach is needed.

\begin{figure}[htb]
\centering 
\includegraphics[width=0.9\linewidth]{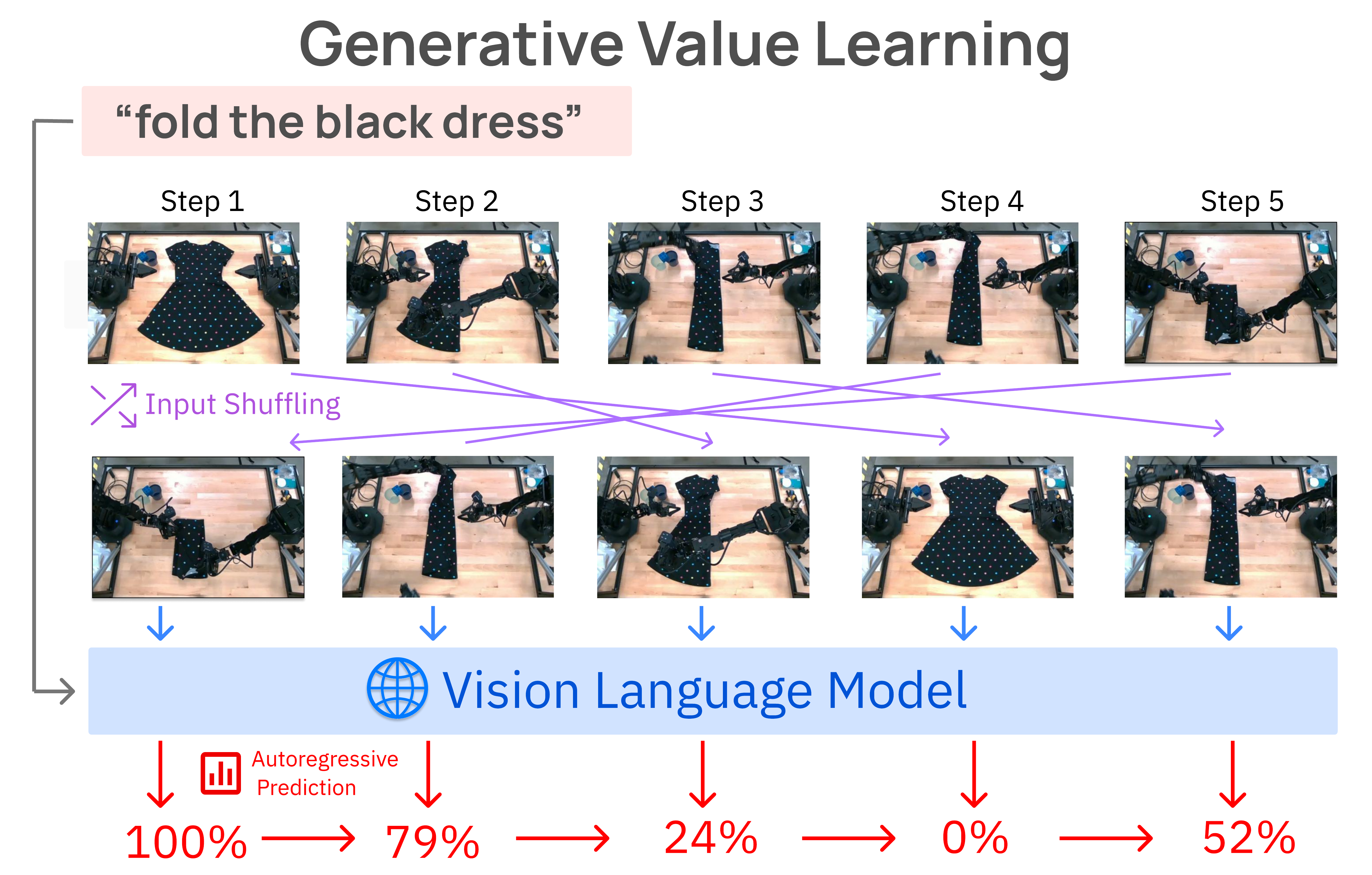}
\caption{\textbf{Method overview.} Generative Value Learning (\ourmethod) generates values by auto-regressively predicting task completion percentage over shuffled frames, enabling impressive in-context value learning.}
\label{fig:concept}
\end{figure}

To effectively leverage the broad knowledge of VLMs, we introduce Generative Value Learning (\ourmethod), a universal value estimation method enabled by long-context VLMs, which crucially operates over \textit{shuffled} frames. At its core, \ourmethod asks frozen state-of-the-art VLMs, such as \texttt{Gemini-1.5-Pro}~\citep{reid2024gemini}, to auto-regressively predict the completion percentage of a task specified in natural language for a sequence of shuffled input video frames; see \cref{fig:concept}. Perhaps surprisingly we find that simply shuffling the frames of the input video effectively overcomes the strong implicit temporal bias found in video, enabling VLMs to generate meaningful values. While \ourmethod is capable of generating values in a zero-shot manner, we find that the performance of \ourmethod scales with examples via multi-modal in-context learning. Providing more examples of visual ``unshuffling'' in context increases performance, irrespective of the target embodiment. For example, human videos can improve \ourmethod's performance on predicting robot task progress.

To facilitate large-scale value prediction evaluation, we additionally introduce a new evaluation metric, Value-Order Correlation (VOC), measuring how well predicted values correlate with the ground-truth timestep order in expert videos; as we will show, VOC is also a useful metric for measuring dataset and trajectory quality, which allows \ourmethod to be used for applications beyond value-based policy learning such as data quality estimation and success detection. We first evaluate \ourmethod's value prediction quality with VOC on a large suite of real-world robotics datasets, spanning 51 datasets, 20 embodiments, and more than 300 tasks. This includes 50 datasets from Open-X (OXE) dataset ~\citep{padalkar2023open} in addition to our own bimanual manipulation dataset containing 250 challenging real-world tasks on an ALOHA platform~\citep{zhao2023learning}, which are considerably longer horizon and more fine-grained than those in the OXE dataset. In aggregate, \ourmethod exhibits strong zero-shot value prediction capabilities with highly positive VOC scores on most datasets; its performance further improves with various types of multi-modal in-context examples. Using \ourmethod, we demonstrate scalable foundation model supervision for robot learning at various data abstraction levels. Specifically, \ourmethod can help measure dataset quality in OXE. Second, it can be used for success detection, enabling imitation learning on mixed-quality datasets. Finally, the raw value estimates from \ourmethod can be used for advantage-weighted regression for real-world offline reinforcement learning~\citep{peters2007reinforcement,peng2019advantage}.

In summary, our contributions are 
\begin{enumerate}[nosep]
    \item Generative Value Learning (\ourmethod), a universal value prediction framework via VLM in-context autoregressive value estimation on shuffled video frames.
    \item An extensive evaluation on real-world datasets demonstrating \ourmethod's zero-shot scalability and multi-modal in-context learning capabilities.
    \item Demonstration that \ourmethod can be used in downstream applications including dataset quality estimation, success detection, and advantage-weighted regression for real-world control. 
\end{enumerate}

\section{Related Work} 
\textbf{Reward and value foundation models.}
Several works have tried to learn transferable reward and value functions from diverse data. Early works learned models using robot \citep{sermanet2016unsupervised} or even human videos with discriminators~\citep{chen2021learning}, contrastive learning \citep{baumli2023vision} or offline RL \citep{ma2022vip, ma2023liv,bhateja2023robotic} to guide manipulation tasks. With the advent of recent language and vision foundation models, several works have integrated them into various robotic applications such as semantic planning \citep{ahn2022can,huang2023grounded,singh2023progprompt,zhang2023bootstrap,ding2023task}, imitation learning ~\citep{brohan2023rt, szot2023large}, and symbolic programming \citep{tang2023saytap, liang2023code, singh2023progprompt, wang2023demo2code, huang2023instruct2act, liu2023llm+, silver2023generalized, ding2023task, lin2023text2motion, xie2023translating}. Most related to our work, LLMs and VLMs have been used as reward models. \citet{kwon2023language,mahmoudieh2022zero} use language models to provide reward values for RL agents, while \citet{klissarov2023motif,wang2024rl,kwon2023language}  use them to provide preference feedback. \citet{ma2023eureka,yu2023language,xie2023text2reward} even have LLMs generate their code. These works use only the language capabilities of foundation models. More recent works directly use VLMs as zero-shot reward models \citep{rocamonde2023vision} or success detectors \citep{du2023vision,guan2024tasksuccess}. Critically, in these works the VLM acts only as an (often sparse) reward function which predicts success, and not a \emph{value} function that predicts task progress. Though some works use chain-of-thought prompting \citep{venuto2024code} or active learning \citep{kwon2023toward}, they generally do not make use of the autoregressive, long-context, or in-context learning capabilities of state-of-art VLMs. As a consequence, they often evalaute \emph{reward} prediction only on simple and simulated tasks. To our knowledge, we are the first to demonstrate that VLMs are capable of generalizable per-frame value estimation on real world tasks which can be used for downstream tasks like dataset selection. \looseness=-1

\textbf{In-context learning for robotics.} In-context learning has been explored in the robot learning literature, primarily focusing on action generation~\citep{duan2017one,finn2017one,dasari2021transformers, xu2022prompting,di2024keypoint, liang2024learning, fu2024context}. However, all these prior works require explicit, and often extensive training, on their robot tasks in order to realize in-context learning capabilities, and generalization is achieved only on narrow distribution of tasks. In contrast, we demonstrate that visual value estimation already enjoys flexible multi-modal in-context learning from pre-trained VLMs without any robot specific fine-tuning.

\section{Generative Value Learning}

In this section, we introduce Generative Value Learning, \ourmethod. At a high level, \ourmethod frames value estimation as an autoregressive next-token prediction problem in which a VLM is tasked with outputting the task progress for a \textit{batch of shuffled trajectory frames}.

\textbf{Problem setup.}  We model robotics tasks as goal-conditioned partially observed Markov decision processes~\citep{puterman2014markov}: $\mathcal{M}(\phi) := (O, A, R, P, T \mu, G)$ with observation space $O$, action space $A$, reward function $R$, transition function $P$ , task horizon $T$, initial state distribution $\mu(o)$, and goal space $G$ that specifies the task semantically. Conditioned on a task $g$ an agent $\pi: O \rightarrow A$ aims to maximizes its value function, or the expected cumulative reward over the task horizon, $V^\pi(o_1;g) = \mathbb{E}_{\mu,\pi, P}[r(o_1;g) + \dots + r(o_T;g)]$. However, reward and value functions can be difficult to define for robotics applications given their heterogeneity. Given this, a popular universal notion of value is task progress~\citep{sermanet2016unsupervised,sermanet2018time,eysenbach2020c,tian2020model,lee2021generalizable}. This kind of temporal value function maps an observation and goal specification to a real number between $0$ and $1$: $V: \mathcal{O} \times \mathcal{G} \rightarrow [0, 1]$, where initial observations of the environment have value $0$ and goal-satisfying observations have value $1$. Under this definition, an expert trajectory $\tau=(o_1,\dots,o_T) \sim \pi_E$, has value function $V^{\pi_E}(o_t;g) = \frac{t}{T}$. In this work, our goal is to obtain such a temporal value function $V$ that can predict such task progress $v_1, \dots v_T$ for each frame of video $o_1,\dots,o_T$.  \looseness=-1

Though we seek to leverage priors imbued in large foundation models, as shown in~\cref{sec:experiments} simply prompting a VLM with video frames fails to produce meaningful estimates. To make VLMs amenable to value prediction, we propose three key components that comprise the \ourmethod method: 1) autoregressive value prediction, 2)input observation shuffling, and 3) in-context value learning.

\textbf{1. Autoregressive value prediction.} 
Traditionally, value functions $V(\cdot): \mathcal{O} \rightarrow \mathbb{R}$ are trained to be self-consistent by enforcing the bellman equation
\begin{equation}
    \label{eq:feed-forward-value}
    V^\pi(o_t) = R(o_t) + \mathbb{E}_{\pi, P}\left[V(o_{t+1})\right].
\end{equation}
When parameterizing a value function as a feed-forward neural network, this is typically done by minimizing the mean-squared error of the equality above. As values for different observations within the same trajectory are related via the bellman equation, the resulting value function remains consistent even if we query it with only a single observation. VLMs on the other hand are not inherently trained with any consistency objective. Thus, if we independently query a VLM with different observations from the same trajectory it is likely to produce inconsistent values.
Our insight is that providing the entire trajectory as input instead of just a single observation offers VLMs greater opportunity to generate self-consistent value estimates. Concretely, given a language description of the task $l_\text{task}$ we ask the VLM to auto-regressively generate values given the entire video as context:
\begin{equation}
    v_{t} = \vlm(o_{1},\dots,o_{T}; v_{1},\dots,v_{t-1};l_{\text{task}}), \forall t \in [2, T].
\end{equation}
We abbreviate this auto-regressive prediction process as $v_{1},\dots,v_{T} = \vlm(o_{1},\dots, o_{T};l_{\text{task}})$. This simple mechanism allows the VLM to attend to all previous predictions and frames when making the next value prediction, enabling it to produce globally consistent estimates over long-horizon sequences without needing to be trained like classical feed-forward value functions. Though this design choice enables VLMs to produce consistent values, it doesn't necessitate that the values are meaningfully. Nai\"vely prompting a VLM in this manner tends to produce linear, monotonic value functions for every single video, regardless of optimality. 

\textbf{2. Input observation shuffling.} Empirically we find that when presented a choronological sequence of frames VLMs discover the short-cut solution of outputting monotonically increasing values, often ignoring the task description or the actual quality of the trajectory. One hypothesis is that as VLMs are trained on ordered video frames for captioning and question answering, the chronology itself is a cue for downstream tasks unrelated to value prediction. As a consequence, model nai\"ve prompting results in unfaithful low-quality value predictions.
To break this temporal bias, we propose randomly shuffling the input frames. In this manner, \ourmethod forces the VLM to pay attention to each individual frame and output faithful value predictions using all information provided in context. Concretely, \ourmethod prompts a VLM as:
\begin{equation}
    v_{\tilde{1}},\dots,v_{\tilde{T}} = \vlm(o_{\tilde{1}}, \dots, o_{\tilde{T}};l_{\text{task}}, o_1), \quad  \text{where}
    \quad  ({\tilde{1}}, \dots, {\tilde{T}}) = \texttt{permute}(1,\dots,T).
\end{equation}
where the permute operator randomly shuffles the temporal indicies. Note however, that we cannot shuffle \textit{every} frame. If we do so, then the arrow of time in the original video can be ambiguous -- i.e., in many cases, the reverse video is also physically plausible, making is the ground-truth order impossible to predict. Thus, as in the above equation we condition the VLM on the first input frame allow it to use the first observation as an anchor point for all other shuffled frames.

\textbf{3. In-context value learning.} 
While auto-regressive prediction and shuffling are enough to obtain good performance, \ourmethod can perform even better by leveraging the appealing properties of VLMs. Notably, large models often exhibit in-context learning, where tasks can be learned by simply providing examples~\citep{brown2020language}. This enables flexible and versatile in context value learning, by which \ourmethod's predictions can steadily improve by providing examples at test time without any model fine-tuning. In particular, we can simply prepend shuffled videos and their ground-truth task progress as in-context examples to boost the value prediction quality via few-shot learning:
\begin{equation}
\label{eq:icl}
    v_{\tilde{1}},\dots,v_{\tilde{T}} = \vlm\left(o_{\tilde{1}},\dots, o_{\tilde{T}}, l_{\text{task}} \mid \texttt{permute}\left((o_1,v_1),(o_2,v_2),\dots, (o_M, v_M)\right)\right)
\end{equation}
As we show in~\cref{sec:experiments}, \ourmethod benefits from flexible forms of in-context examples, including videos from unrelated tasks and even humans. Though \ourmethod zero-shot is already effective across a broad range of tasks and robots, in-context learning can still realize substantial improvement on the most difficult bimanual dexterous tasks.

\textbf{Practical implementation.}
To predict temporal value functions in practice,  \ourmethod asks the VLM to output integer-valued percentage numbers between 0 and 100. Given that real-world robot video datasets are of different lengths and taken at different frequencies, we subsample all videos so that there are 30 frames in the input sequence to ensure comparable findings across datasets.  See the Appendix for the full prompt and implementation.

\section{Experiments}
\label{sec:experiments}

We conduct large scale experiments assessing \ourmethod's value prediction generalization and in-context learning capabilities. Specifically, we study the following questions:

\begin{enumerate}[nosep]
\item Can \ourmethod produce zero-shot value predictions for a broad range of tasks and embodiments?
\item Can \ourmethod improve from in-context learning? 
\item Can \ourmethod be used for other downstream robot learning applications?
\end{enumerate}

In all our experiments, we use \texttt{Gemini-1.5-Pro}~\citep{reid2024gemini} as the backbone VLM for \ourmethod; we ablate this model choice and find \ourmethod effective with other VLMs as well. After thorough study of \ourmethod's value prediction capabilities, we study several downstream applications in visuomotor policy learning, aiming to improve data quality at dataset, trajectory, and individual transition levels.

\noindent \textbf{Evaluation metric.} 
Our goal is to evaluate \ourmethod value estimation at scale on as many robot datasets as possible, holistically testing its generalization capabilities and understanding its limitations. This makes it difficult to use traditional evaluation metrics for value functions, such as observing downstream learned policy performance, as they require value functions that are specifically trained or finetuned for individual tasks and embodiments. This quickly becomes very expensive for universal value functions that are intended for use across a large set of diverse real-world tasks and robots, many of which the practitioner may not have access to. Prior works on large-scale value learning have resorted to visually observing the smoothness of the value curve on expert trajectories as a qualitative ``eye-test'' for model generalization~\citep{ma2022vip, ma2023liv,karamcheti2023language}, but such evaluation is conducted on only few selected videos. We formalize and scale up this intuitive approach and introduce
a lightweight, yet predictive method for evaluating value models: Value-Order Correlation (VOC). This metric computes the \textit{rank correlation} between the predicted values and the chronological order of the input expert video:
\begin{equation}
    \label{eq:correlation-metric}
    \texttt{VOC} = \texttt{rank-correlation}\left(\texttt{argsort}(v_{\tilde{1}}, \dots, v_{\tilde{T}}); \texttt{arange}(T) \right); 
\end{equation}
VOC ranges from $-1$ to $1$, where $1$ indicates that the two orderings are perfectly aligned. Expert quality demonstrations, by construction, have values that monotonically increase with time, and thus a good value model should have high VOC scores when evaluated on expert videos. On the other hand, fixing a good value model, low-quality trajectories should have low VOC scores. This is because sub-optimal trajectories often contain high repetition of visually similar frames due to the presence of redundant, re-attempt actions or poorly-placed cameras. As such, the values along the trajectories should not be monotonic, resulting in low correlation with the ground-truth timestep order. As we will show in our experiments, this value rank correlation metric has strong predictive power for the quality of the values as well as downstream policy learning performance, validating its usefulness as a standalone evaluation metric for value predictions.

\subsection{Large-scale real-world evaluation}
\label{sec:oxe}
To study \ourmethod's zero-shot value prediction capability, we evaluate its VOC on two large expert robotics datasets.

\textbf{Open X-Embodiment dataset.} First, we consider the Open X-Embodiment (OXE) dataset~\citep{padalkar2023open}. an aggregation of trajectory data from 50 standalone academic robot datasets that consists of diverse tasks, robots, and camera viewpoints. For each of the 50 datasets, we randomly sample 20 trajectories and evaluate \ourmethod zero-shot on each of the sampled trajectories. Note that not all OXE datasets have language task annotations, so we use the last frame of the trajectory as the goal specification when text annotation is not provided. To better contextualize \ourmethod's value prediction quality, we compare to a state-of-the-art multi-modal value model \textbf{LIV}~\citep{ma2023liv}, a contrastive vision-language model~\citep{radford2021learning} fine-tuned with value learning objective on human videos for in-the-wild value estimation. LIV predicts the temporal value of an input observation by computing its embedding distance to the embedding of the goal image or task description. 

\begin{figure*}[t!]
    \centering
    \includegraphics[width=0.49\textwidth]{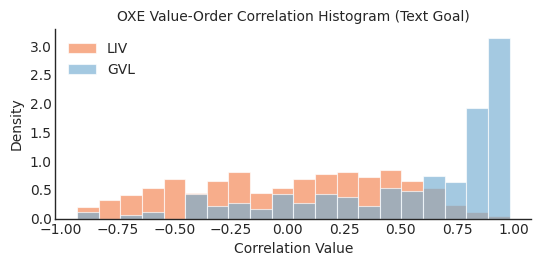}
    \includegraphics[width=0.49\textwidth]{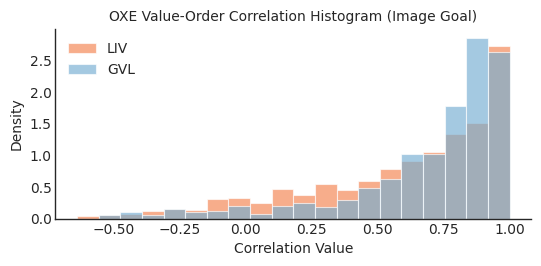}
    \caption{\textbf{Zero-shot value predictions on OXE datasets.} Left: \ourmethod significantly outperforms LIV on datasets with language goals. Right: \ourmethod still outperforms LIV on datasets with image goals despite solving the more difficult task of frame re-shuffling.}
    \label{fig:oxe_sweep}
\end{figure*}

For evaluation, we plot the histogram of all 1000 (50$\times$20) Value Order Correlation (VOC) scores in \cref{fig:oxe_sweep}, split by goal modalities. Given that most OXE datasets contain human-collected expert demonstrations, good value models should have high VOC scores; however, we acknowledge that there are sub-optimal trajectories within OXE that can introduce noise in our results. After we first establish \ourmethod as an effective universal value model. we will present how \ourmethod can be used to detect low-quality data in~\cref{sec:applications}. As shown in~\cref{fig:oxe_sweep}, on both goal modalities, \ourmethod consistently generates VOC scores that heavily skew to the right, indicating that it is able to zero-shot recover the temporal structure hidden in the shuffled demonstration videos, i.e., coherent value predictions. \ourmethod's performance is also markedly better than LIV on language goals (\cref{fig:oxe_sweep} left). Here, LIV's predictions are random, suggesting that its embedding space does not contain sufficient knowledge for predicting dense values for arbitrary unseen robot videos. On image goals, LIV's prediction problem is arguably simpler because an embedding space that simply captures image similarity can result in ascending values that correlate with timesteps. Even then, \ourmethod generates better quality value predictions as judged by slightly higher VOCs  (\cref{fig:oxe_sweep} right).
In summary, \ourmethod can indeed effectively utilize the world knowledge afforded by the backbone VLM to achieve effective value predictions zero-shot for the breadth of real-world robotic tasks and datasets.

\begin{figure}[h]
\centering 
\includegraphics[width=0.49\textwidth]{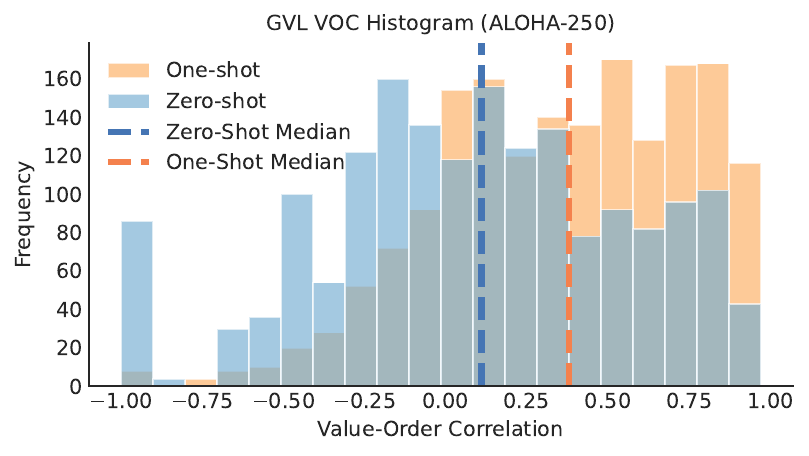}
\includegraphics[width=0.49\textwidth]{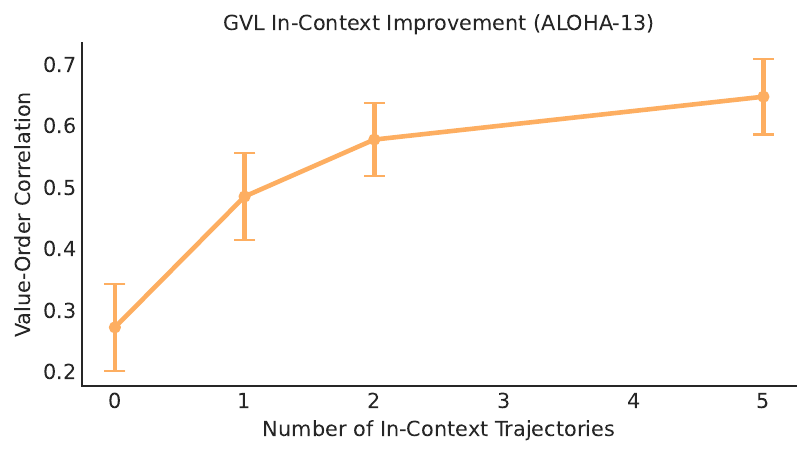}
\caption{\ourmethod scales up to 250 ALOHA bi-manual tasks and can improve with in-context examples.}
\label{fig:aloha_sweep}
\end{figure}

\begin{figure}[htbp]
    \begin{subfigure}[b]{0.49\textwidth}
        \centering
        \includegraphics[width=\textwidth]{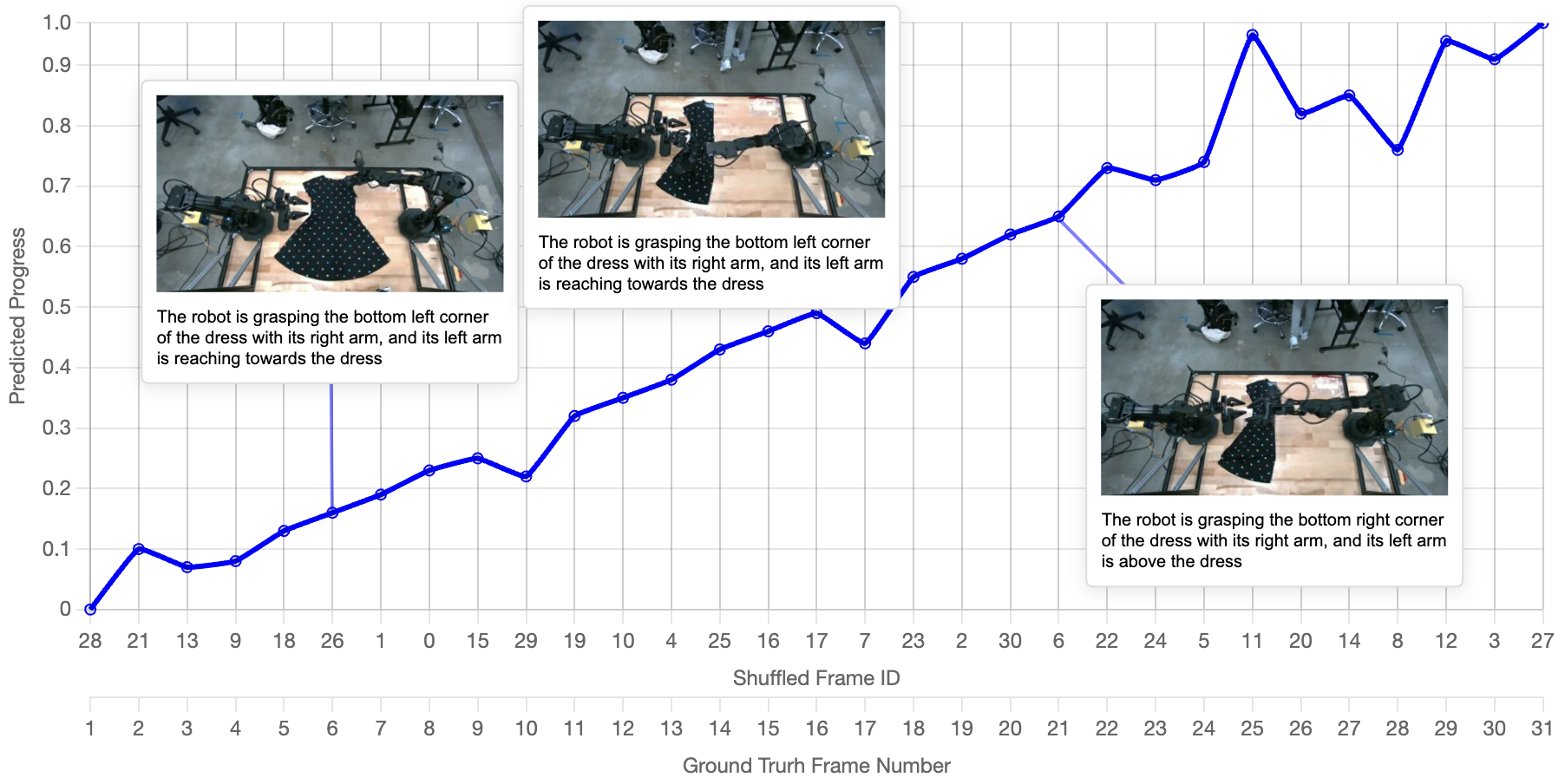}
        \caption{\texttt{fold\_dress} from the top-down view.}
        \label{fig:fold_dress_vis}
    \end{subfigure}
    \hfill
    \begin{subfigure}[b]{0.49\textwidth}
        \centering
        \includegraphics[width=\textwidth]{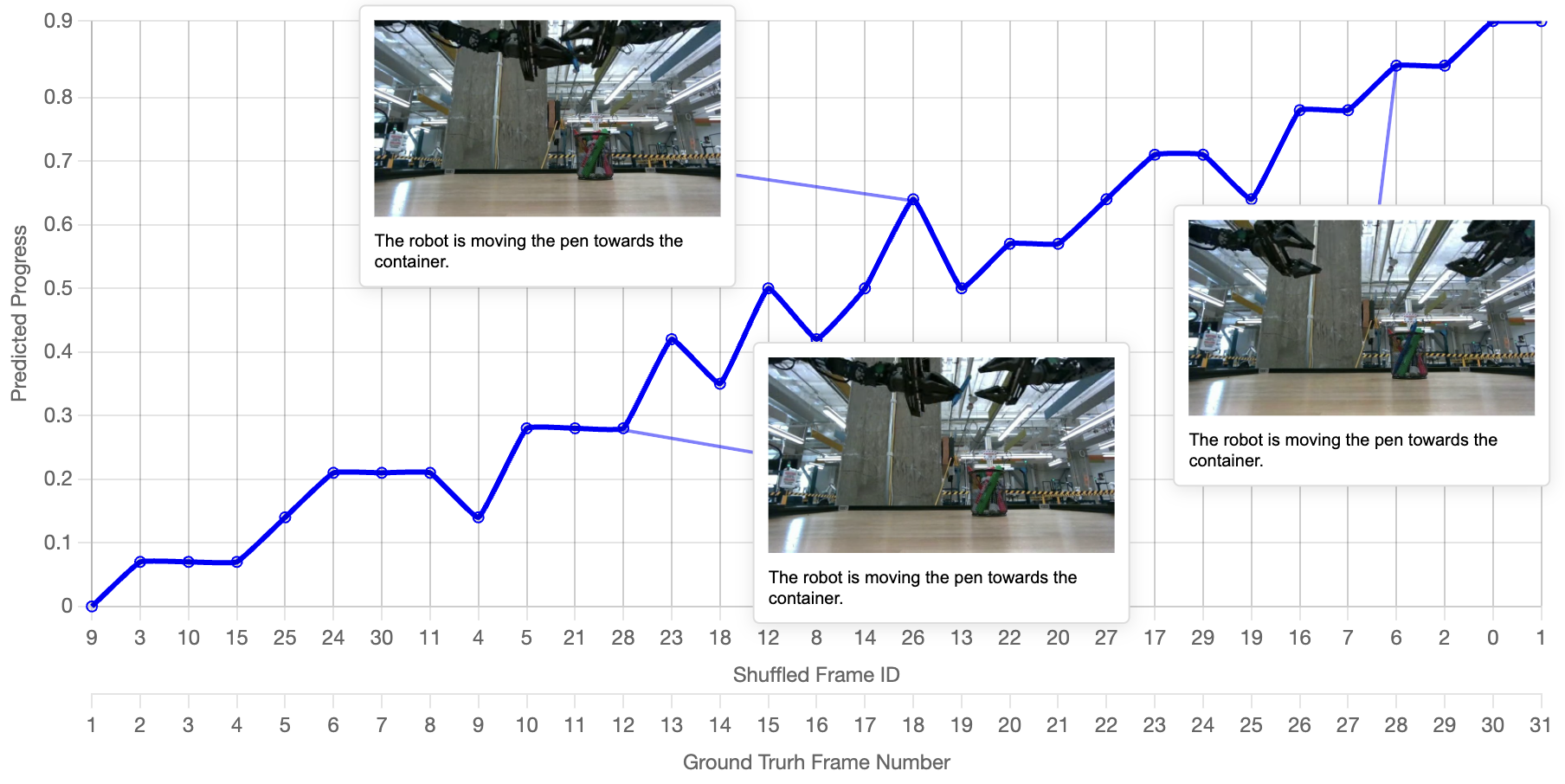}
        \caption{\texttt{pen\_handover} from the table view.}
        \label{fig:pen_handover_vis}
    \end{subfigure}
    \vskip\baselineskip
    \begin{subfigure}[b]{0.49\textwidth}
        \centering
        \includegraphics[width=\textwidth]{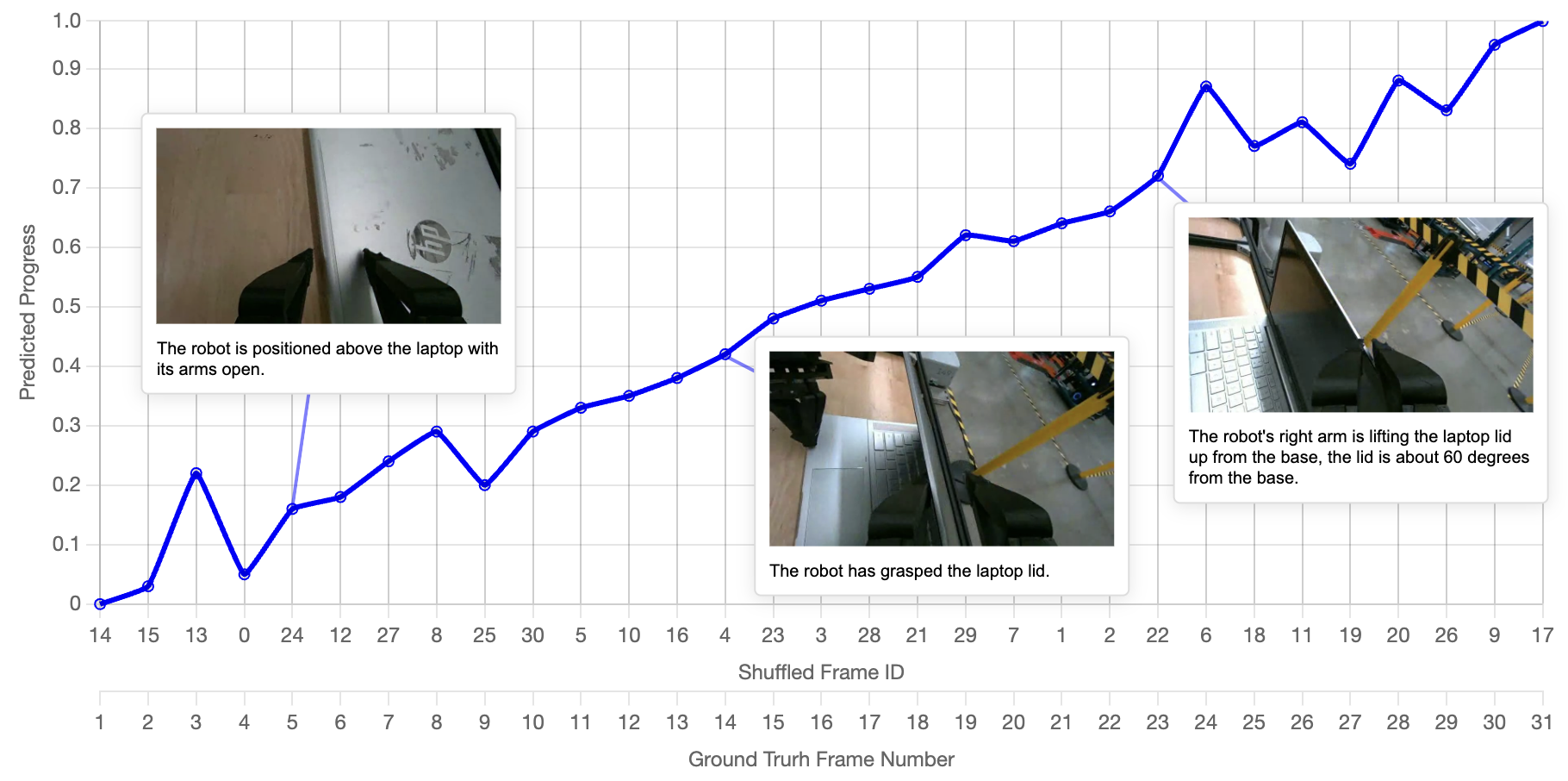}
        \caption{\texttt{close\_laptop} from the right wrist camera.}
        \label{fig:close_laptop_vis}
    \end{subfigure}
    \hfill
    \begin{subfigure}[b]{0.49\textwidth}
        \centering
        \includegraphics[width=\textwidth]{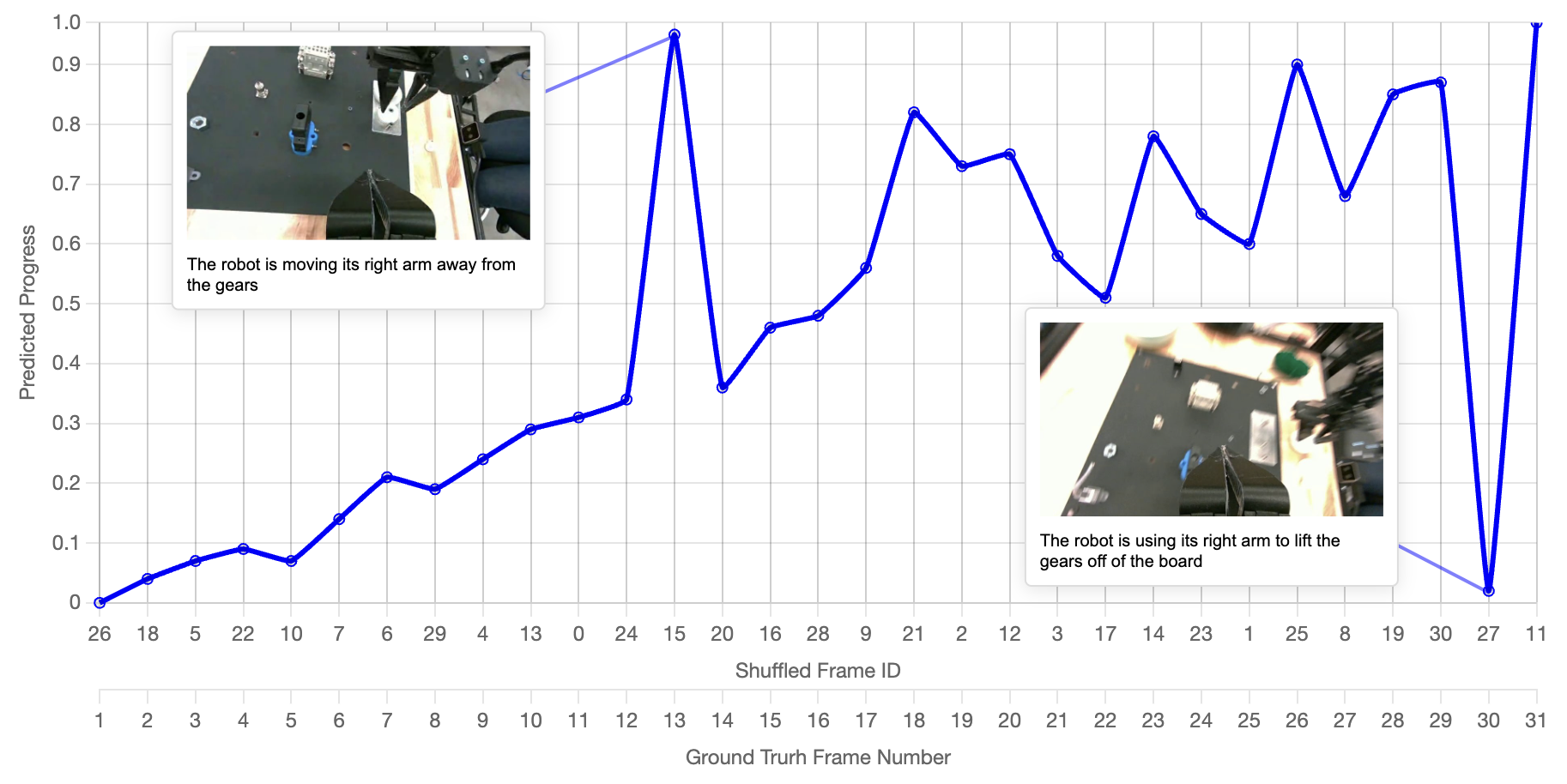}
        \caption{\texttt{remove\_gears} from the left wrist camera.}
        \label{fig:remove_gears_vis}
    \end{subfigure}
\caption{Example \ourmethod predictions on real-world ALOHA tasks. \ourmethod can successfully unshuffle video frames and generate meaningful task values on diverse tasks and camera viewpoints.}
    \label{fig:aloha_vis}
\end{figure}

\textbf{Challenging bimanual datasets.} 
OXE datasets primarily focus on simpler, short horizon single-arm tasks. 
To further stress test \ourmethod, we evaluate on a new diverse dataset of 250 distinct household tabletop tasks on the bi-manual ALOHA systems~\citep{zhao2023learning,aldaco2024aloha}. This dataset includes highly challenging, long-horizon skills, such as removing three gears sequentially from a NIST board, folding a dress in eighth-fold, hanging a t-shirt on a cloth rack. See the bottom right of \cref{fig:concept} for representative ALOHA tasks.
For each task, we evaluate on 2 human teleoperated demonstrations to evaluate \ourmethod zero-shot. The aggregate histogram over all 500 (250 $\times$ 2) VOC scores is illustrated in \cref{fig:aloha_sweep}. As shown, \ourmethod is capable of generating value predictions that are positively correlated on more than 60\% of them with median VOCs of 0.12. This is promising, but worse than the performance on the OXE datasets. While our main results use the top-down camera shown in \cref{fig:icv_cross_embodiment}, we show the effect of using \ourmethod across each of the four cameras used in our ALOHA setup in \cref{fig:icv_camera_ablations}. In \cref{fig:aloha_vis}, we present several qualitative examples of GVL predictions; see our \href{https://generative-value-learning.github.io/}{project website} for additional examples. In the next section, we extensively explore how to improve \ourmethod on this dataset using in-context learning techniques.

\subsection{Multi-Modal In-Context Value Learning } 
As the diverse ALOHA dataset is significantly more challenging, we explore whether \ourmethod can benefit from in-context learning, where additional shuffled observation-value pairs are presented in the VLM context window (Eq.~\ref{eq:icl}). 

\textbf{Few-shot in-context learning.} First, we collect an additional demonstration for each of the 250 tasks and use its shuffled value-observation pairs as context for one-shot \ourmethod value prediction for the same set of 500 evaluations. As seen in \cref{fig:aloha_sweep}, with one in-context trajectory, \ourmethod's performance substantially improves with 90\% positive VOCs and a median VOC of 0.37. We further investigate whether performance can improve with more in-context examples on a represented subset of 13 tasks for which have more than $500$ demonstrations. For these tasks, we evaluate few-shot \ourmethod on 500 distinct trajectories per task with up to 5 in-context examples.  The average VOCs over tasks and trajectories is shown in \cref{fig:aloha_sweep} (Right). We see that \ourmethod demonstrates appealing in-context scaling as the average score steadily improves as we increase the number of in-context examples. Even with 5 in-context trajectories, meaning 150 total shuffled images, \ourmethod is able to utilize its full context and exhibit strong generalization. This result demonstrates how state-of-art long-context-window VLMs, such as \texttt{Gemini-1.5-Pro}, can be re-purposed to make for general-purpose value functions with impressive test-time improvement capability, quickly mastering value predictions with minimal supervision. \looseness=-1

\begin{figure*}
    \centering
    \begin{minipage}[b]{0.6\textwidth}
        \begin{subfigure}[t]{\textwidth}
            \centering
            \includegraphics[width=0.3\textwidth]{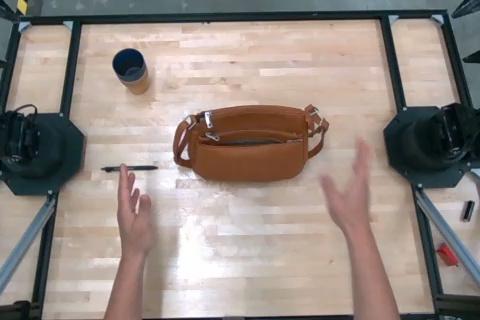}
            \hfill
            \includegraphics[width=0.3\textwidth]{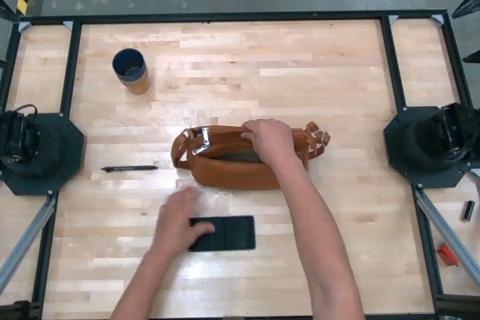}
            \hfill
            \includegraphics[width=0.3\textwidth]{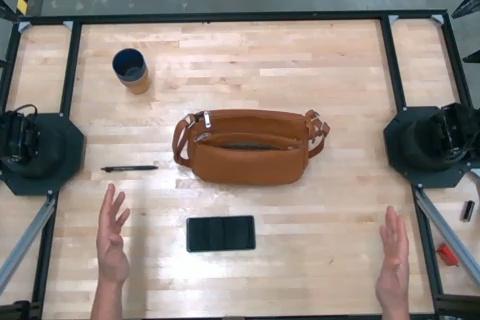}
            \caption{\small{In-context examples from human videos}}
        \end{subfigure}
        \begin{subfigure}[t]{\textwidth}
            \centering
            \includegraphics[width=0.3\textwidth]{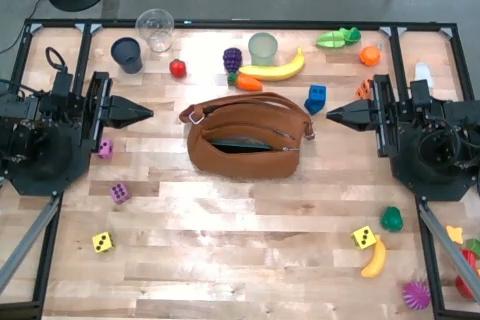}
            \hfill
            \includegraphics[width=0.3\textwidth]{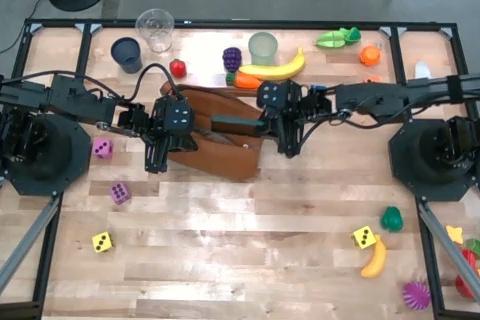}
            \hfill
            \includegraphics[width=0.3\textwidth]{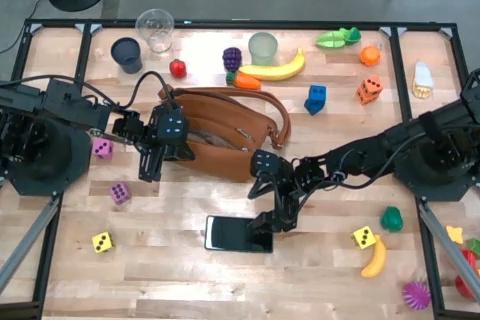}
            \caption{\small{Target prediction robot videos}}
        \end{subfigure}
    \end{minipage}
    \hfill
    \begin{minipage}[b]{0.39\textwidth}
        \centering
        \includegraphics[width=\textwidth]{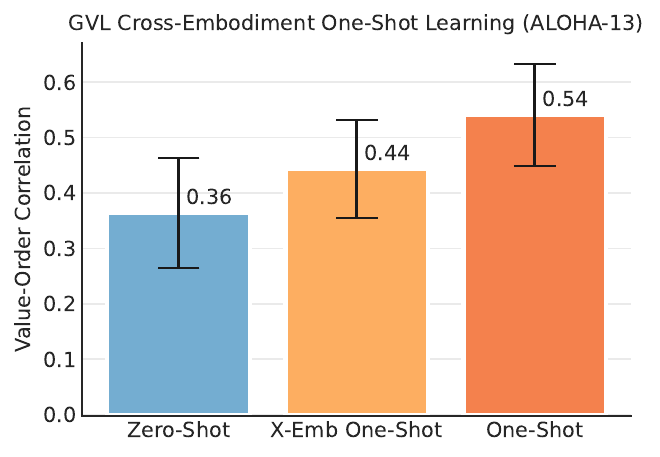}
    \end{minipage}
    \caption{\ourmethod benefits from cross-embodiment in-context learning capability: its value predictions can be improved by examples from human videos.}
    \label{fig:icv_cross_embodiment}
\end{figure*}

\textbf{Cross-embodiment in-context learning.}
Examples in-context are not limited to robot demonstrations. One advantage of \ourmethod is that it can still benefit from in-context learning even when the demonstrations come from a different embodiment. Specifically, we record humans performing the same tasks as the ALOHA robot demonstrations  and then use these human demonstrations as in-context examples for value prediction. As shown in \cref{fig:icv_cross_embodiment}, \ourmethod with one cross-embodiment in-context example can effectively improve over its zero-shot counterpart. In the Appendix, we also show that \ourmethod can similarly benefit from \textit{cross-task} in-context learning. In conclusion, \ourmethod presents a versatile framework for in-context value learning that can scale up to even the most challenging manipulation tasks.

\subsection{\ourmethod Applications}

\label{sec:applications}
As \ourmethod can generate high-quality value estimates, it can be applied to a number of downstream tasks including dataset quality estimation, success detection, and weighted imitation learning.

\begin{wraptable}{r}{0.5\textwidth}
\vspace*{-1em}
\centering
\resizebox{0.3\textwidth}{!}{
\begin{tabular}[c]{l|c}\toprule
Dataset & Avg. VOC \\ \midrule
RT-1~\citep{brohan2022rt} & 0.74 \\ 
Dobb-E~\citep{shafiullah2023bringing} & 0.53 \\ 
Bridge~\citep{walke2023bridgedata} & 0.51 \\ 
QT-OPT~\citep{kalashnikov2018scalable} & 0.19 \\ 
DROID~\citep{khazatsky2024droid} & -0.01 \\ 
RoboNet~\citep{dasari2019robonet} & -0.85 \\ 
\bottomrule
\end{tabular}}
\caption{Average Value-Order Correlation (VOC) scores on selected OXE datasets. As shown, demonstration datasets with un-occluded camera views generally have high scores. In contrast, exploration datasets have low scores.}
\label{table:icv_oxe_subset}
\end{wraptable}
\textbf{Dataset Quality Estimation.} Robotic action models are increasingly trained on large mixtures of datasets~\citep{padalkar2023open,team2024octo,kim2024openvla} and selecting the right mixture  is critical for policy performance \citep{hejna2024remix}. However, dataset mixing is often done in an ad-hoc fashion by visual inspection \citep{team2024octo}. Having validated that \ourmethod is an effective zero-shot value model, we investigate whether we can in turn use \ourmethod's VOC scores to determine dataset quality within OXE. 
To this end, for each OXE dataset in \cref{fig:oxe_sweep}, we compute the average correlation scores for its sampled trajectories and present the ranking of the average score in~\cref{appendix:icv_oxe_breakdown}. In~\cref{table:icv_oxe_subset}, we present a subset of selected representative large-scale datasets in OXE. We see that datasets have large spread in their VOC scores, but these scores are interpretable and match human intuitions. Specifically, datasets collected from human teleoperators with relative fixed camera placements, such as RT-1~\citep{brohan2022rt}, Dobb-E~\citep{shafiullah2023bringing}, and Bridge~\citep{ebert2021bridge,walke2023bridgedata}, have high VOC scores, despite their diversity in scenes and tasks. In contrast, datasets with autonomous data collection via scripted motions or motor babbling, such as QT-OPT~\citep{kalashnikov2018scalable} and RoboNet~\citep{dasari2019robonet}, contain high number of suboptimal trajectories that do not exhibit smooth temporal structure to be re-shuffled.

Interestingly, DROID~\citep{khazatsky2024droid}, a recent large household manipulation  ataset is ranked very low, consistent with prior works ~\citep{kim2024openvla} that found that removing DROID from large action model training improved final performance. After inspecting trajectories from DROID with a low VOC score from \ourmethod we found that many have poor camera angles that do not capture robot motion or have the arm or manipulated objects heavily occluded. These observations indicate that \ourmethod VOC can be indicative of dataset quality.

\begin{wraptable}{r}{0.5\textwidth}
\resizebox{0.5\textwidth}{!}{
\begin{tabular}[c]{lccc}
\toprule
Method & Accuracy & Precision & Recall \\
\midrule
\ourmethod-SD (Zero-Shot) & 0.71 & 0.71 & 0.71 \\
\midrule
\ourmethod-SD (One-Shot) & \textbf{0.75} & \textbf{0.85} & 0.70 \\
\midrule
SuccessVQA~\citep{du2023vision} & 0.62 & 0.33 & \textbf{0.73} \\
SuccessVQA-CoT & 0.63 & 0.44 & 0.68\\
\bottomrule 
\end{tabular}}
\caption{Comparison of VLM success detectors.}
\label{table:sd_comparison_sim}
\end{wraptable} 

\textbf{Success detection and filtered imitation learning.} Next we consider more granular intra-dataset quality control by investigating how \ourmethod can be used as a success detector for trajectory filtering, enabling filtered imitation learning on mixed quality datasets. As discussed, good value models should return low VOC scores on unsuccessful trajectories; in particular, it is difficult for \ourmethod to re-shuffle frames within sub-optimal trajectories which often contain irregular or repetitive behavior.
Thus, we can use \ourmethod for success detection by filtering trajectories that have VOC scores below certain numerical threshold; we refer to this procedure as \ourmethod-SD. We evaluate \ourmethod-SD on six simulated bimanual dexterous manipulation tasks on the ALOHA system (see \cref{fig:simulation-tasks}). Simulation is well-suited for this experiment because we can naturally control for data quality and reproducibility. More specifically, for each task, we construct a mixed quality dataset by rolling out a pre-trained policy of roughly 50\% success rate for 1000 episodes, mirroring real-world autonomous data collection settings with high failure rate~\citep{kalashnikov2018scalable}. We compare to \textbf{SuccessVQA}~\citep{du2023vision}, which poses success detection as a Visual-Question Answering problem. To ensure that the same amount of information is provided, we feed the full video sequence to the VLM; therefore, this baseline tests whether the VLM is equipped with video understanding capability good enough for out-of-the-box success detection. In addition, we consider \textbf{SuccessVQA-CoT}, which uses chain-of-thought prompting~\citep{wei2022chain} to encourage the VLM to output intermediate textual reasoning outputs before providing the final success answer. Unless otherwise stated we use a VOC threshold of $0.5$ for \ourmethod-SD. \looseness=-1

\begin{figure}[h]
\centering 
\begin{subfigure}[b]{0.49\textwidth}
    \includegraphics[width=\textwidth]{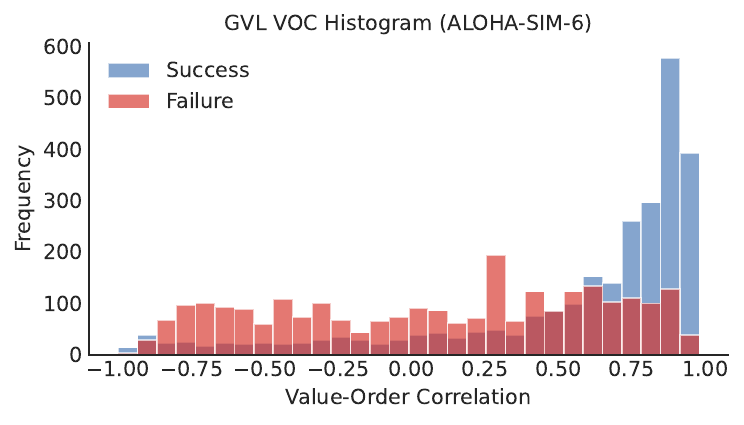}
    \label{fig:icv_sim_histogram_aggregate}
\end{subfigure}
\hfill
\begin{subfigure}[b]{0.49\textwidth}
    \includegraphics[width=\textwidth]{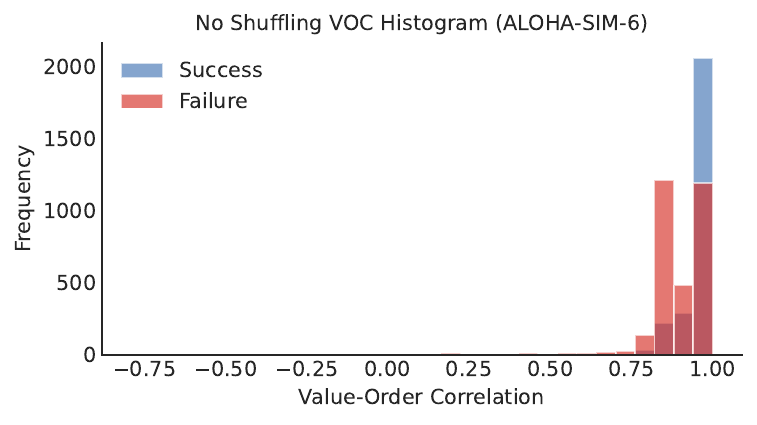}
    \label{fig:icv_noshuffling_sim_histogram_aggregate}
\end{subfigure}

\caption{\textbf{\ourmethod for Success Detection.} Left: \ourmethod behaves qualitatively differently on successful and failed trajectories. Right: \ourmethod (No-Shuffling) loses discriminability on failure trajectories. }
\label{fig:icv_sd_histogram}
\end{figure}

\begin{wraptable}{r}{0.5\textwidth}
\centering
\resizebox{0.5\textwidth}{!}{
\begin{tabular}{lccc}
\toprule
Real-World ALOHA Tasks & \ourmethod + DP & DP & \begin{tabular}[c]{@{}c@{}} Avg. VOC\end{tabular} \\
\midrule
\texttt{bowl-in-rack} & \textbf{7/10} & 6/10 & 0.57 \\
\texttt{banana-handover} & \textbf{7/10} & 5/10 & 0.73 \\
\texttt{close-laptop} & \textbf{9/10} & 6.5/10 & 0.59 \\
\texttt{open-drawer} & 4/10 & \textbf{6/10} & 0.09 \\
\texttt{remove-gears} & 4.67/10 & \textbf{7/10} & 0.19 \\
\texttt{pen-handover} & \textbf{1.5/10} & 0/10 & 0.43 \\
\texttt{fold-dress} & \textbf{7/10} & \textbf{7/10} & 0.66 \\
\bottomrule 
\end{tabular}}
\caption{\textbf{Real-World ALOHA Policy Learning Results.} AWR with \ourmethod (One-Shot) outperforms IL baselines when the predicted values have high VOCs.}
\label{tab:icv-aloha-realworld}
\end{wraptable}

For all methods, we report the accuracy, precision, and recall in ~\cref{table:sd_comparison_sim}. \ourmethod-SD consistently outperforms or matches SuccessVQA  on all classification metrics. In particular, SuccessVQA has low precision, indicating that the base VLM systematically biases towards outputting failure. Adding one in-context demonstration further improves \ourmethod's performance across all metrics. In \cref{fig:icv_sd_histogram} (Left), we also visualize the histogram of the VOC scores \ourmethod produces on success and failure trajectories. As expected, \ourmethod on failure trajectories renders a uniform distribution when the task is unsuccessful, indicating the model's inability to uncover the original temporal order -- success and failure trajectories have distinct distributions over the correlation values indicating that \ourmethod can adequately separate them. \cref{fig:icv_sd_histogram} (Right) shows that the histograms without shuffling are largely the same independent of success or failure. This shows that by forcing the VLM to perform the more difficult prediction task over shuffled frames, \ourmethod can elicit better zero-shot values. 

\begin{figure*}[t!]
    \centering
    \includegraphics[width=0.68\textwidth]{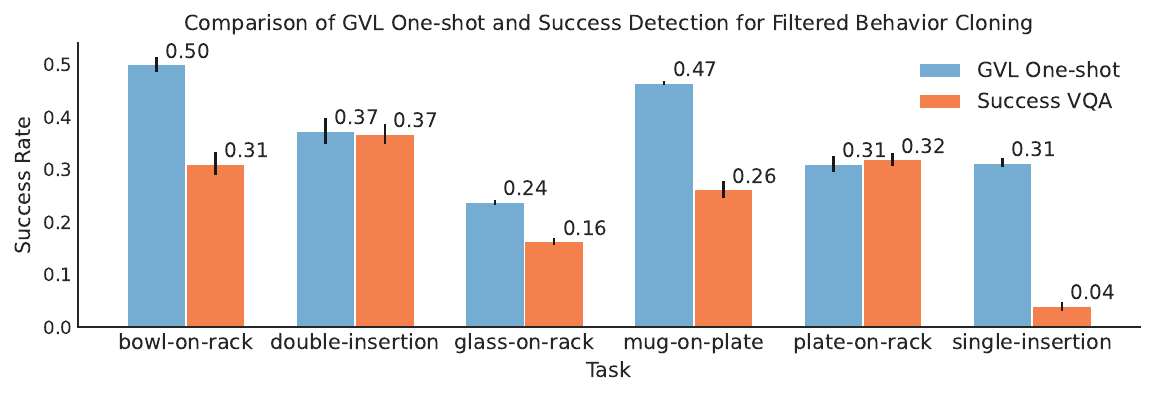}
    \includegraphics[width=0.31\textwidth]{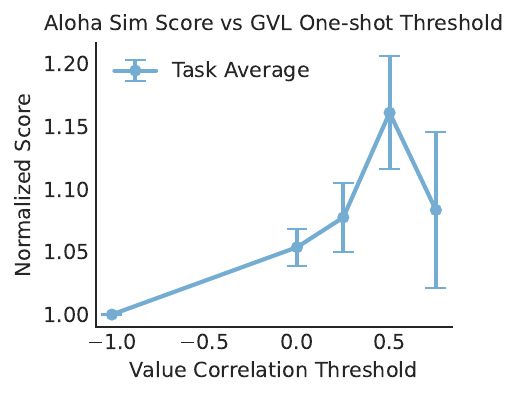}
    \caption{\textbf{Success-Filtered Imitation Learning on ALOHA Simulation Tasks.} Left: Using \ourmethod-SD for success-filtered BC substantially outperforms SuccessVQA. Right: \ourmethod-SD is not sensitive to the VOC threshold for improving imitation learning.}
    \label{fig:icv_filtered_bc}
\end{figure*}

Now, we use the above success detection methods for filtered imitation learning; for all methods, we use Action Chunking Transformer (ACT) as the imitation learning algorithm~\citep{zhao2023learning}; ACT hyperparameters are tuned for ACT on the success-only subset and are fixed for all methods. Given the noisiness in model checkpoints performance, we report the average success rate of the last 10 model training checkpoints. Results for the six simulation tasks are shown to the right of \cref{fig:icv_filtered_bc}, where \ourmethod-SD's improved success detection leads to better performance over SuccessVQA. In fact, SuccessVQA often hurts performance, likely because of its low precision which causes the policy to train on a high number of false positive (i.e. failure) trajectories. In \cref{fig:icv_filtered_bc} (Right) we show the effect of varying the VOC threshold in $\{-1.0, 0, 0.25, 0.5, 0.75\}$ in comparison to training on all the data with ACT; note that this is the same using the lowest threshold, $-1.0$ as it is a lower bound on the VOC metric. As seen, \ourmethod consistently outperforms ACT regardless of threshold values; when the threshold value is too high, i.e., $0.75$, we see a slight dip in performance when the overall dataset size becomes too small.

\begin{figure*}[t!]
    \centering
    \begin{subfigure}[b]{0.24\textwidth}
        \includegraphics[width=\textwidth]{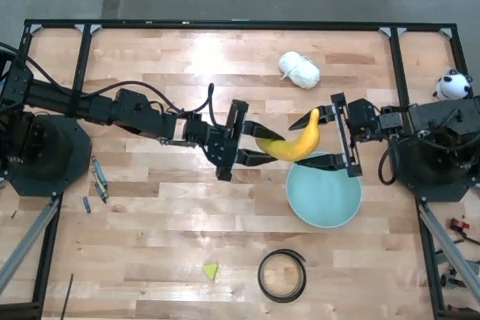}
        \caption{\texttt{banana-handover}}
    \end{subfigure}
    \hfill
    \begin{subfigure}[b]{0.24\textwidth}
        \includegraphics[width=\textwidth]{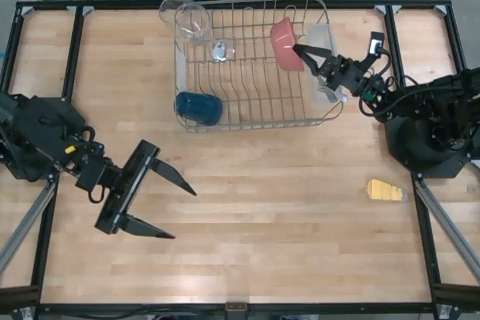}
        \caption{\texttt{bowl-in-rack}}
    \end{subfigure}
    \hfill
    \begin{subfigure}[b]{0.24\textwidth}
        \includegraphics[width=\textwidth]{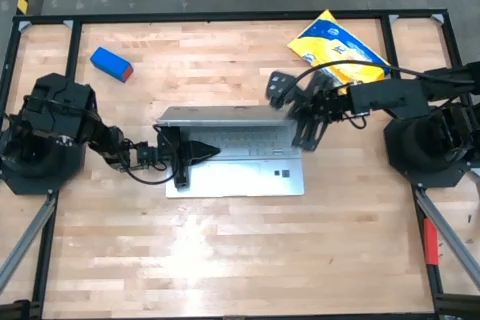}
        \caption{\texttt{close-laptop}}
    \end{subfigure}
    \newline 
    \begin{subfigure}[b]{0.24\textwidth}
        \includegraphics[width=\textwidth]{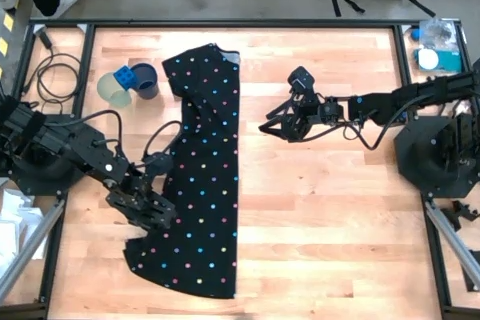}
        \caption{\texttt{fold-dress}}
    \end{subfigure}
    \hfill
    \begin{subfigure}[b]{0.24\textwidth}
        \includegraphics[width=\textwidth]{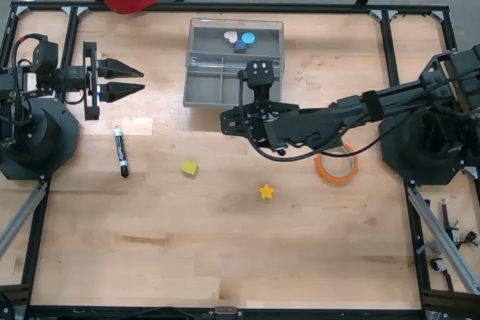}
        \caption{\texttt{open-drawer}}
    \end{subfigure}
    \hfill
    \begin{subfigure}[b]{0.24\textwidth}
        \includegraphics[width=\textwidth]{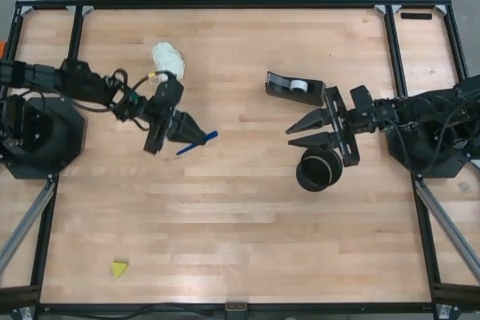}
        \caption{\texttt{pen-handover}}
    \end{subfigure}
    \hfill
    \begin{subfigure}[b]{0.24\textwidth}
        \includegraphics[width=\textwidth]{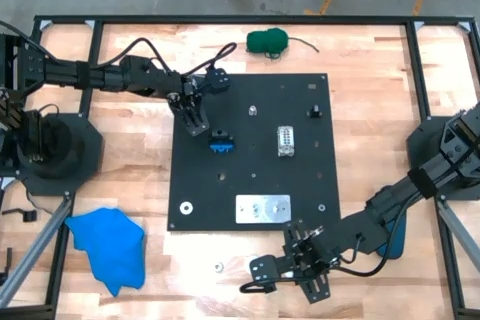}
        \caption{\texttt{remove-gears}}
    \end{subfigure}
    
    \caption{Real-world ALOHA tasks for real-word policy evaluation.}
    \label{fig:real-tasks}
\end{figure*} 

\textbf{Advantage-weighted regression for real-world visuomotor control.}
Finally, we illustrate how \ourmethod can assign importance weights to individual transitions within trajectories at a fine-grained level akin to offline reinforcement learning. For these experiments we use real-world demonstration data collected by human teleoperation on bi-manual ALOHA robot setups. Unlike simulation, our datasets only contain successful task executions but can be sub-optimal and multi-modal. Thus, we directly utilize \ourmethod's values with \textit{advantage weighted regression } (AWR)~\citep{peters2007reinforcement, peng2019advantage}, in which we weight each individual transition by the estimated advantange, or \ourmethod value difference for that step:  
\begin{equation}
        \label{eq:awr} 
    \mathcal{L}(\theta) := - \mathbb{E}\left[  \mathrm{exp}\left( \tau \cdot (v_{k+1}-v_{k}) \right) \cdot \log \pi_\theta(a_k \mid o_k)\right]
\end{equation}

If an action were deemed to make progress towards the task goal, then its future value $v_{k+1}$ should be significantly higher than the present value $v_k$, leading to a large positive weight. By upweighting the most promising actions in diverse human-collected datasets, we posit that AWR can outperform imitation learning approaches.

We use diffusion policy (DP) as the policy backbone~\citep{chi2023diffusion} for each task, and compare training diffusion policies with \ourmethod (One-Shot) advantage weighting or lack thereof. We evaluate on 7 tasks with 10 trials per task and report success rate in~\cref{tab:icv-aloha-realworld}. As can be seen, on a majority tasks, \ourmethod-DP outperforms DP and we see a clear correlation between improvement over DP and the VOC score. That is, when the value predictions are of high quality as judged by VOC, policy learning can benefit from \ourmethod value weighting. On \texttt{open-drawer} and \texttt{remove-gears}, the top-down view does not provide sufficient resolution to distinguish task progress (see~\cref{fig:real-tasks}), as a consequence, the value predictions can be noisy, which can hurt policy learning. However, given the in-context learning results, we believe that it is possible to improve policy learning even on difficult tasks with non-ideal camera viewpoints.

\subsection{Ablations}

Finally, we ablate key algorithmic design choices of \ourmethod to validate their necessity. In the Appendix, we additionally demonstrate that \ourmethod's performance is robust to the choice of backbone VLMs as well as input 
camera viewpoint. 

\textbf{Is autoregressive value prediction necessary?} 
We consider an ablation that simply asks the VLM to predict values of input observations one by one without \ourmethod's autoregressive batch prediction mechanism. This ablation, which we refer to as \textbf{VLM (Single Frame)}, essentially poses value estimation as a VQA problem. We compare this ablation to \ourmethod on a subset of RT-1 dataset as in~\cref{sec:oxe}; the average VOC for VLM (Single Frame) is a mere $-0.08$, a significant drop from \ourmethod's 0.74 on RT-1 dataset. As seen, pre-trained VLMs by themselves are poor value estimators, generating inconsistent values that are too noisy to be used in practice.

\textbf{Is input observation shuffling necessary?}
As discussed, we find that removing shuffling collapses ICV's predictions into generating degenerate values; that is, regardless of the quality of the provided trajectory, \ourmethod tends to predict monotonically increasing values, resulting in inflated VOC scores that cannot be used to discriminate successful and failure trajectories.; see~\cref{fig:icv_sd_histogram} (Right). To further qualitatively illustrate this phenomenon, in~\cref{fig:icv_shuffle_ablations} in the Appendix, we overlay raw \ourmethod value predictions with frame shuffling and lackthereof to understand the spread of the value curves. We see that the overlay for original \ourmethod looks ``messy'', suggesting that \ourmethod outputs varied value curves that better capture the heterogeneity of the queried video qualities. In contrast, without frame shuffling, \ourmethod predictions indeed collapses onto a few linear ascending patterns.

\section{Conclusion}
We have introduced Generative Value Learning (\ourmethod), a universal value function via VLM autoregressive value prediction on shuffled video frames.~\ourmethod can zero-shot output dense and high-quality value predictions for diverse and challenging real-world robotic tasks, spanning various robot embodiments and task categories. With few-shot learning from the same task, different task, or different embodiment, \ourmethod performance steadily improves. We have demonstrated several use cases of using \ourmethod to perform dataset, trajectory, and transition selection to improve downstream policy learning performance and generalization. We believe that \ourmethod takes an important step in using foundation models supervision for robot learning. 

\textbf{Limitations and future work.} We have not investigated whether pre-trained VLMs can be fine-tuned to perform better value predictions. In addition, though we test on diverse camera viewpoints, we have not yet investigated whether multi-view observations can improve value prediction quality. In addition, our evaluation metric Value-Order Correlation is most suitable for a-periodic tasks for which there exists a unique ordering of frames from an expert demonstration. Tasks such as wiping or stirring may be hard to discern. Though these limitations present avenues for future work, we believe \ourmethod is a step towards improved in-the-wild value estimation.

\section*{Acknowledment}
We thank Jie Tan, Pannag Sanketi, Oliver Groth, and the rest the Google DeepMind Robotics team for helpful discussions and providing feedback on the paper.

\newpage

\bibliography{bibliography}
\bibliographystyle{iclr2025_conference}

\appendix
\newpage 

\section{Prompt}
In this section, we provide the full prompt provided to the VLM for \ourmethod predictions. The same prompt is used for all OXE datasets.

\definecolor{lightgray}{rgb}{0.95,0.95,0.95}

\begin{mdframed}[
    linecolor=gray,
    linewidth=0.5pt,
    roundcorner=10pt,
    backgroundcolor=lightgray,
    innerleftmargin=5pt,
    innerrightmargin=5pt,
    innertopmargin=5pt,
    innerbottommargin=5pt
]
\begin{lstlisting}[
    basicstyle=\ttfamily,
    columns=fullflexible,
    breaklines=true,
    keepspaces=true,
    showstringspaces=false,
    backgroundcolor=\color{lightgray}
]
You are an expert roboticist tasked to predict task completion percentages for frames of a robot for the task of {task_description}. The task completion percentages are between 0 and 100, where 100 corresponds to full task completion. We provide several examples of the robot performing the task at various stages and their corresponding task completion percentages. Note that these frames are in random order, so please pay attention to the individual frames when reasoning about task completion percentage. 

Initial robot scene: [IMG]

In the initial robot scene, the task completion percentage is 0. 

 Now, for the task of {task_description}, output the task completion percentage for the following frames that are presented in random order. For each frame, format your response as follow: Frame {i}: Frame Description: {}, Task Completion Percentages:{}%

Frame 1: [IMG]
...
Frame n: [IMG]

\end{lstlisting}
\end{mdframed}

\section{\ourmethod OXE Dataset VOC Breakdown}
\label{appendix:icv_oxe_breakdown}
In this section, we provide the full list of average VOC score for each OXE dataset. In~\cref{tab:gemini_oxe_scores}, we provide the VOC scores for \ourmethod with \texttt{Gemini-1.5-Pro} as the backbone VLM. In~\cref{tab:gpt_oxe_scores}, we provide the VOC scores for \ourmethod with \texttt{GPT-4o} as the backbone VLM.

\begin{table}
\begin{center}
\resizebox{0.8\textwidth}{!}{
\begin{tabular}{|l|r|}
\toprule
\textbf{Dataset} & \textbf{VOC Score} \\
\midrule
utokyo\_pr2\_opening\_fridge\_converted\_externally\_to\_rlds & 0.8095 \\
utokyo\_xarm\_bimanual\_converted\_externally\_to\_rlds & 0.7955 \\
utokyo\_xarm\_pick\_and\_place\_converted\_externally\_to\_rlds & 0.7880 \\
fractal20220817\_data & 0.7385 \\
maniskill\_dataset\_converted\_externally\_to\_rlds & 0.7260 \\
berkeley\_autolab\_ur5 & 0.7185 \\
nyu\_door\_opening\_surprising\_effectiveness & 0.6685 \\
utokyo\_pr2\_tabletop\_manipulation\_converted\_externally\_to\_rlds & 0.5875 \\
utaustin\_mutex & 0.5810 \\
iamlab\_cmu\_pickup\_insert\_converted\_externally\_to\_rlds & 0.5585 \\
fmb & 0.5555 \\
ucsd\_kitchen\_dataset\_converted\_externally\_to\_rlds & 0.5295 \\
dobbe & 0.5295 \\
toto & 0.5270 \\
bridge & 0.5145 \\
austin\_sirius\_dataset\_converted\_externally\_to\_rlds & 0.5100 \\
asu\_table\_top\_converted\_externally\_to\_rlds & 0.5055 \\
berkeley\_rpt\_converted\_externally\_to\_rlds & 0.4835 \\
berkeley\_cable\_routing & 0.4470 \\
usc\_cloth\_sim\_converted\_externally\_to\_rlds & 0.4410 \\
jaco\_play & 0.4205 \\
bc\_z & 0.4065 \\
viola & 0.4035 \\
berkeley\_mvp\_converted\_externally\_to\_rlds & 0.3900 \\
roboturk & 0.3545 \\
austin\_buds\_dataset\_converted\_externally\_to\_rlds & 0.3415 \\
stanford\_hydra\_dataset\_converted\_externally\_to\_rlds & 0.3325 \\
tokyo\_u\_lsmo\_converted\_externally\_to\_rlds & 0.3140 \\
berkeley\_fanuc\_manipulation & 0.2685 \\
cmu\_stretch & 0.2625 \\
ucsd\_pick\_and\_place\_dataset\_converted\_externally\_to\_rlds & 0.2410 \\
kuka & 0.1915 \\
dlr\_sara\_pour\_converted\_externally\_to\_rlds & 0.1600 \\
taco\_play & 0.0945 \\
dlr\_edan\_shared\_control\_converted\_externally\_to\_rlds & 0.0855 \\
droid & -0.0060 \\
stanford\_robocook\_converted\_externally\_to\_rlds & -0.0690 \\
imperialcollege\_sawyer\_wrist\_cam & -0.1225 \\
kaist\_nonprehensile\_converted\_externally\_to\_rlds & -0.1310 \\
austin\_sailor\_dataset\_converted\_externally\_to\_rlds & -0.1715 \\
cmu\_play\_fusion & -0.3445 \\
stanford\_kuka\_multimodal\_dataset\_converted\_externally\_to\_rlds & -0.3770 \\
stanford\_mask\_vit\_converted\_externally\_to\_rlds & -0.4505 \\
nyu\_franka\_play\_dataset\_converted\_externally\_to\_rlds & -0.4555 \\
uiuc\_d3field & -0.7025 \\
cmu\_franka\_exploration\_dataset\_converted\_externally\_to\_rlds & -0.7395 \\
columbia\_cairlab\_pusht\_real & -0.7625 \\
robo\_net & -0.8485 \\
dlr\_sara\_grid\_clamp\_converted\_externally\_to\_rlds & -1.0000 \\
\hline
\end{tabular}
}
\caption{\ourmethod (\texttt{Gemini-1.5-Pro}) OXE Dataset VOC Scores}
\end{center}
\label{tab:gemini_oxe_scores}
\end{table}

\begin{table}
\centering
\begin{tabular}{|l|r|}
\toprule
\textbf{Dataset} & \textbf{VOC Score} \\
\midrule
nyu\_door\_opening\_surprising\_effectiveness & 0.883 \\
utokyo\_pr2\_opening\_fridge\_converted\_externally\_to\_rlds & 0.864 \\
berkeley\_mvp\_converted\_externally\_to\_rlds & 0.8285 \\
utaustin\_mutex & 0.813 \\
fractal20220817\_data & 0.803 \\
utokyo\_xarm\_pick\_and\_place\_converted\_externally\_to\_rlds & 0.7665 \\
berkeley\_autolab\_ur5 & 0.755 \\
utokyo\_xarm\_bimanual\_converted\_externally\_to\_rlds & 0.749 \\
utokyo\_pr2\_tabletop\_manipulation\_converted\_externally\_to\_rlds & 0.734 \\
austin\_sirius\_dataset\_converted\_externally\_to\_rlds & 0.7235 \\
toto & 0.713 \\
dlr\_edan\_shared\_control\_converted\_externally\_to\_rlds & 0.6595 \\
bridge & 0.6445 \\
berkeley\_fanuc\_manipulation & 0.6295 \\
berkeley\_rpt\_converted\_externally\_to\_rlds & 0.6235 \\
ucsd\_kitchen\_dataset\_converted\_externally\_to\_rlds & 0.603 \\
roboturk & 0.57 \\
jaco\_play & 0.5615 \\
iamlab\_cmu\_pickup\_insert\_converted\_externally\_to\_rlds & 0.557 \\
uiuc\_d3field & 0.5395 \\
usc\_cloth\_sim\_converted\_externally\_to\_rlds & 0.5355 \\
asu\_table\_top\_converted\_externally\_to\_rlds & 0.5025 \\
maniskill\_dataset\_converted\_externally\_to\_rlds & 0.499 \\
kaist\_nonprehensile\_converted\_externally\_to\_rlds & 0.492 \\
viola & 0.4605 \\
austin\_buds\_dataset\_converted\_externally\_to\_rlds & 0.454 \\
cmu\_play\_fusion & 0.4235 \\
tokyo\_u\_lsmo\_converted\_externally\_to\_rlds & 0.3875 \\
austin\_sailor\_dataset\_converted\_externally\_to\_rlds & 0.3015 \\
ucsd\_pick\_and\_place\_dataset\_converted\_externally\_to\_rlds & 0.2675 \\
berkeley\_cable\_routing & 0.255 \\
dlr\_sara\_pour\_converted\_externally\_to\_rlds & 0.252 \\
imperialcollege\_sawyer\_wrist\_cam & 0.239 \\
robo\_net & 0.237 \\
stanford\_hydra\_dataset\_converted\_externally\_to\_rlds & 0.205 \\
cmu\_stretch & 0.1895 \\
bc\_z & 0.176 \\
nyu\_franka\_play\_dataset\_converted\_externally\_to\_rlds & 0.1735 \\
stanford\_robocook\_converted\_externally\_to\_rlds & 0.16 \\
kuka & 0.132 \\
stanford\_mask\_vit\_converted\_externally\_to\_rlds & -0.173 \\
stanford\_kuka\_multimodal\_dataset\_converted\_externally\_to\_rlds & -0.1785 \\
columbia\_cairlab\_pusht\_real & -0.1815 \\
cmu\_franka\_exploration\_dataset\_converted\_externally\_to\_rlds & -0.2075 \\
taco\_play & -0.2705 \\
eth\_agent\_affordances & -0.279 \\
dlr\_sara\_grid\_clamp\_converted\_externally\_to\_rlds & -1 \\
\bottomrule
\end{tabular}
\caption{\ourmethod (\texttt{GPT-4o}) OXE Dataset VOC Scores}
\label{tab:gpt_oxe_scores}
\end{table}

\section{Simulation Tasks}
\begin{figure*}[t!]
    \centering
    \begin{subfigure}[b]{0.3\textwidth}
        \includegraphics[width=\textwidth]{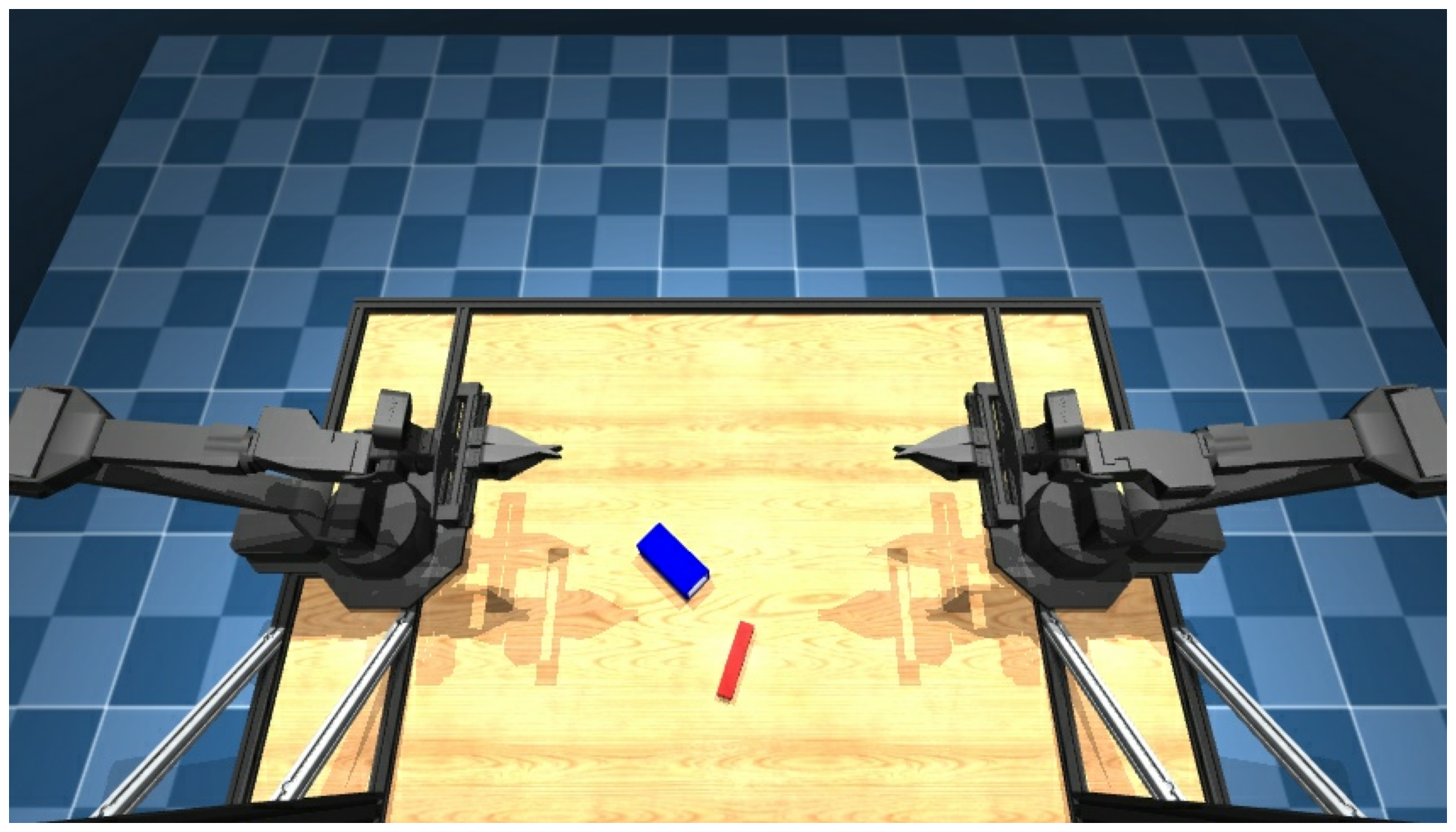}
        \caption{\texttt{single\_insertion}}
    \end{subfigure}
    \hfill
    \begin{subfigure}[b]{0.3\textwidth}
        \includegraphics[width=\textwidth]{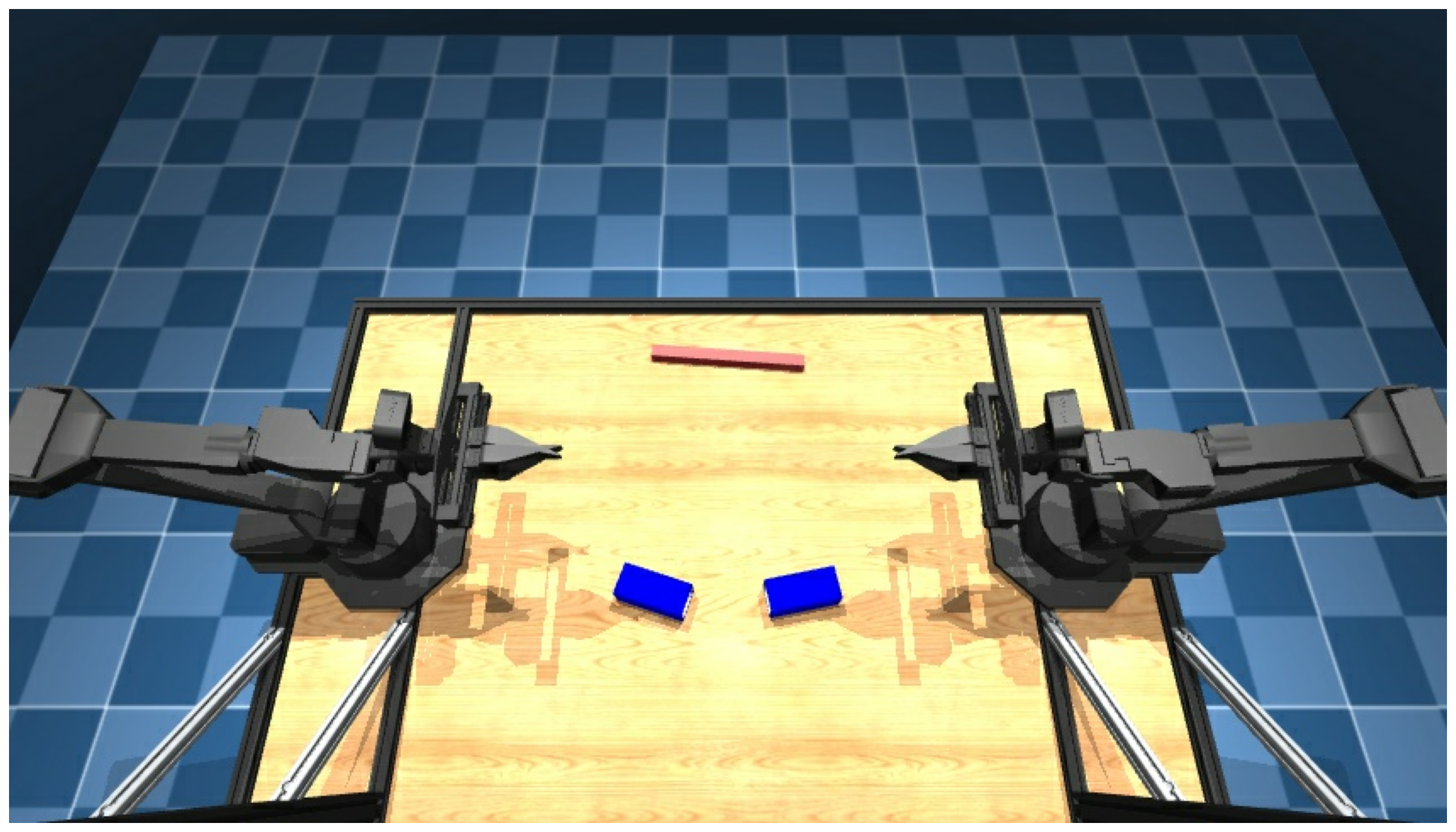}
        \caption{\texttt{double\_insertion}}
    \end{subfigure}
    \hfill
    \begin{subfigure}[b]{0.3\textwidth}
        \includegraphics[width=\textwidth]{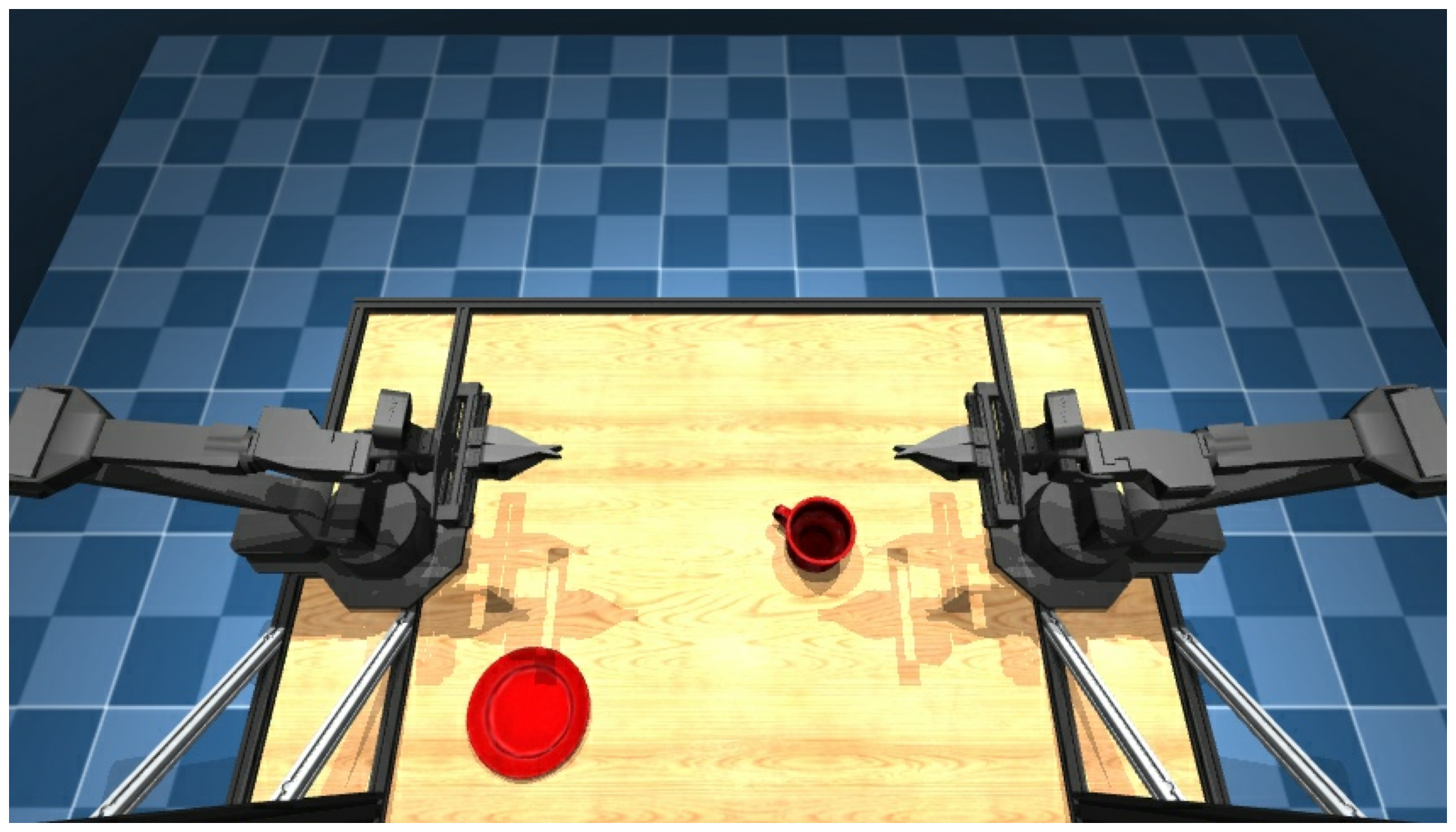}
        \caption{\texttt{mug\_on\_plate}}
    \end{subfigure}
    
    \vspace{1em}
    
    \begin{subfigure}[b]{0.3\textwidth}
        \includegraphics[width=\textwidth]{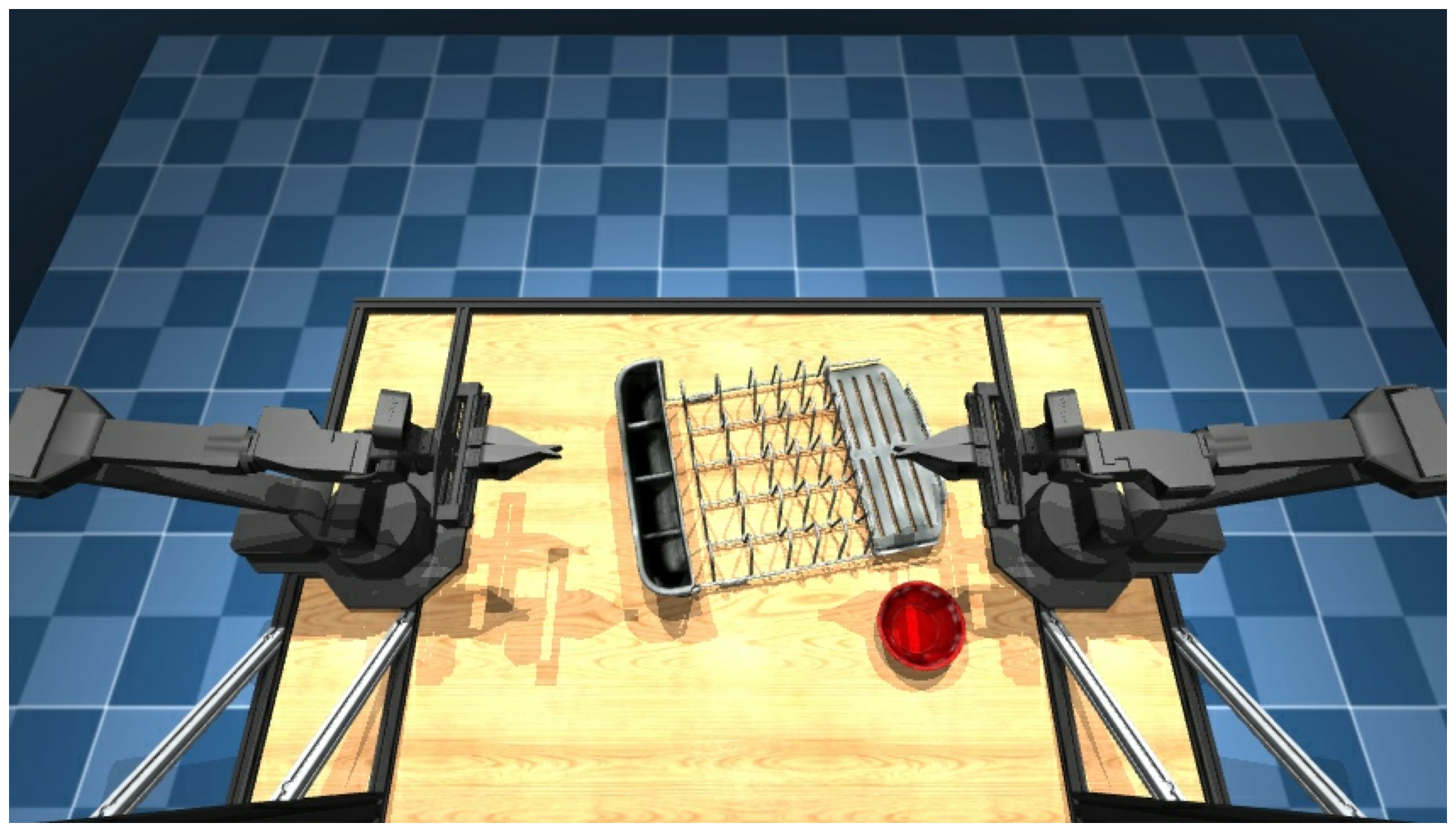}
        \caption{\texttt{bowl\_on\_rack}}
    \end{subfigure}
    \hfill
    \begin{subfigure}[b]{0.3\textwidth}
        \includegraphics[width=\textwidth]{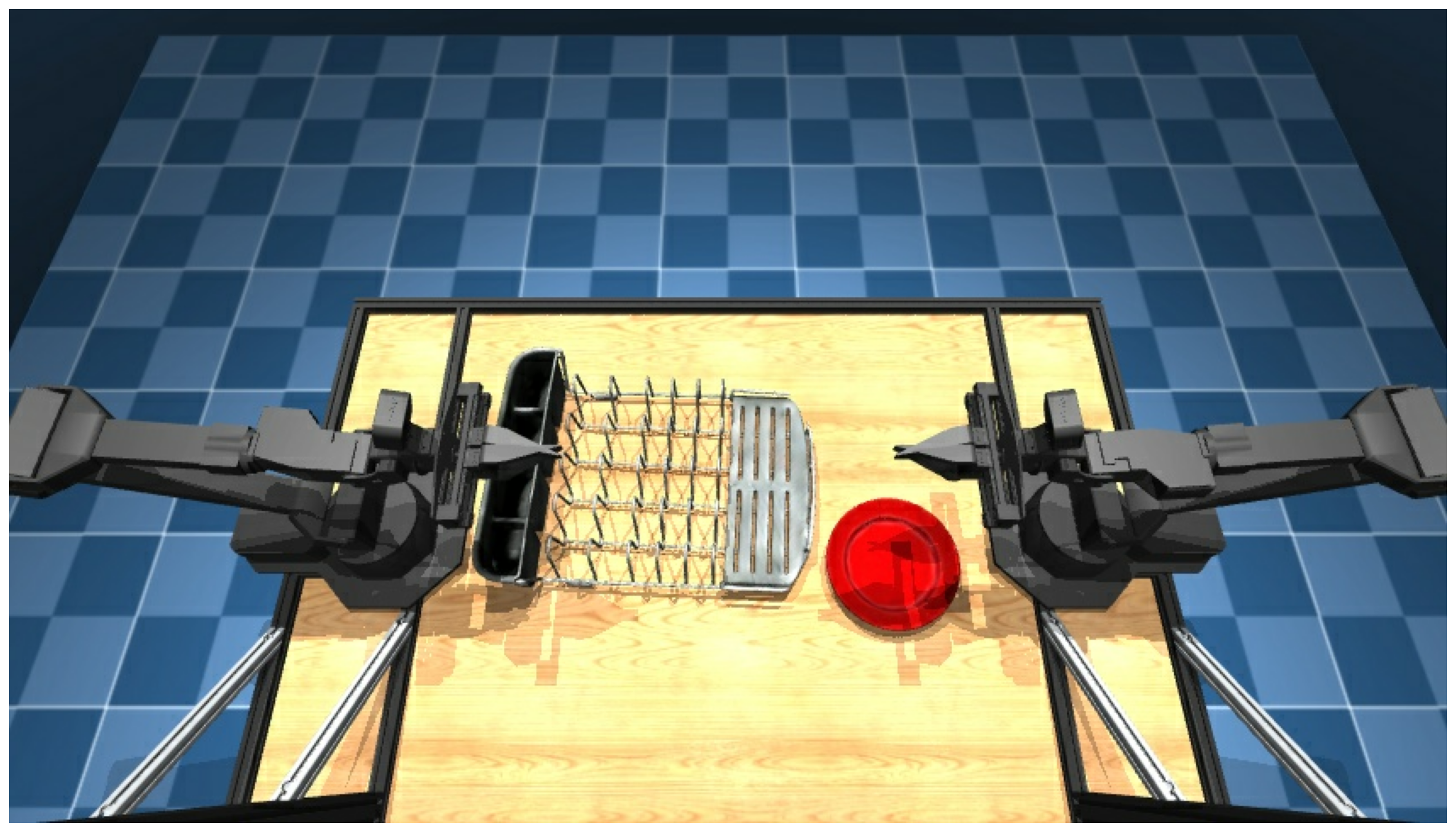}
        \caption{\texttt{plate\_on\_rack}}
    \end{subfigure}
    \hfill
    \begin{subfigure}[b]{0.3\textwidth}
        \includegraphics[width=\textwidth]{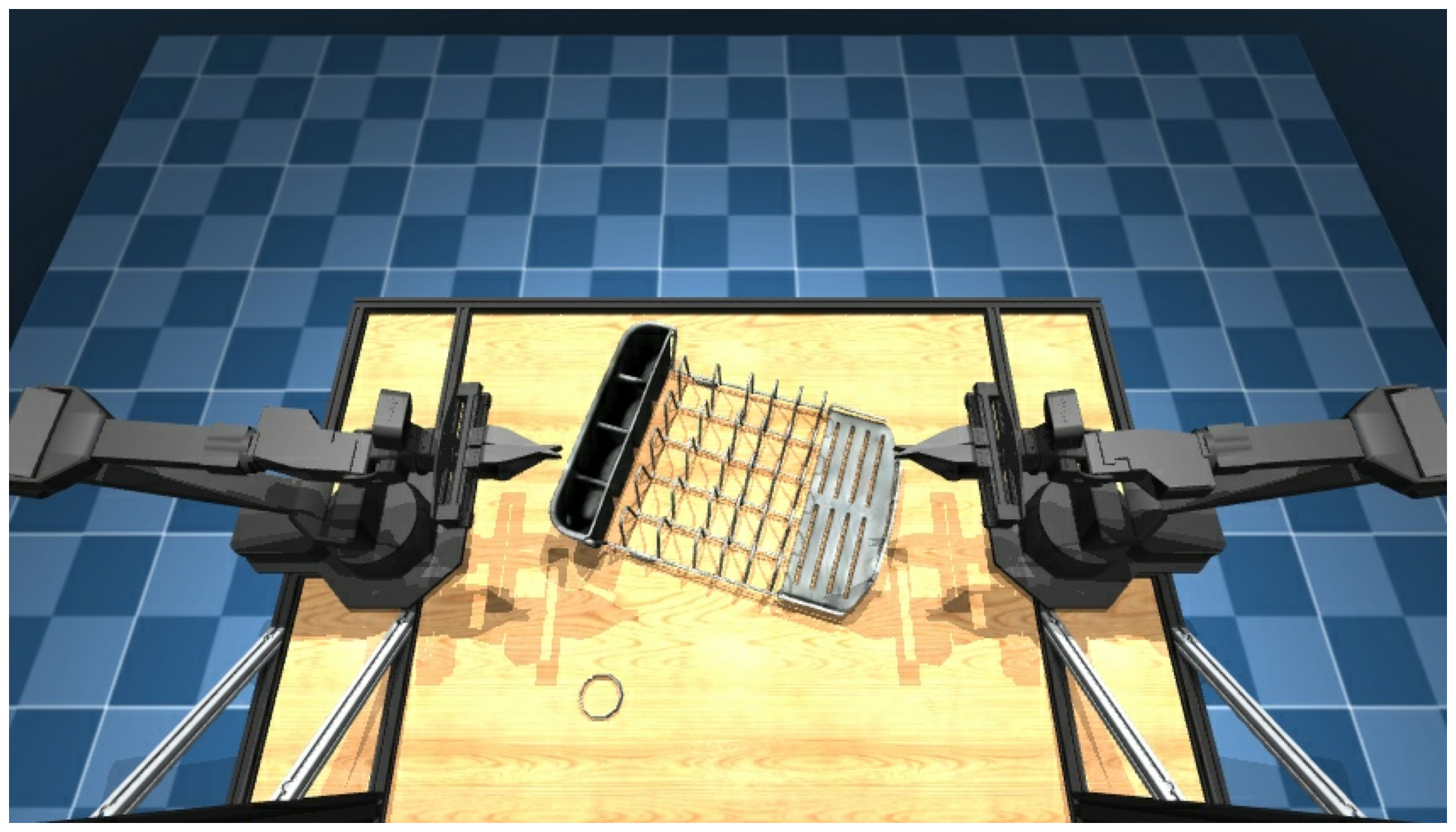}
        \caption{\texttt{glass\_on\_rack}}
    \end{subfigure}
    
    \caption{Simulated task setups of dexterous manipulation on the ALOHA robot.}
    \label{fig:simulation-tasks}
\end{figure*}

In Figure~\ref{fig:simulation-tasks}, we illustrate the six simulation tasks used for the success detection and filtered imitation learning experiment. For each task, we use VR teleoperation to collect 500 trajectories for initial policy training. After the policy converges, we rollout the last checkpoint for 1000 imtes, resulting in naturally balanced mix-quality datasets of about half success and half failure trajectories. 

\section{Additional Results}
\label{appendix:additional-results}

In this section, we present additional results and analysis.

\begin{figure}[h]
\centering 
\includegraphics[width=0.45\textwidth]{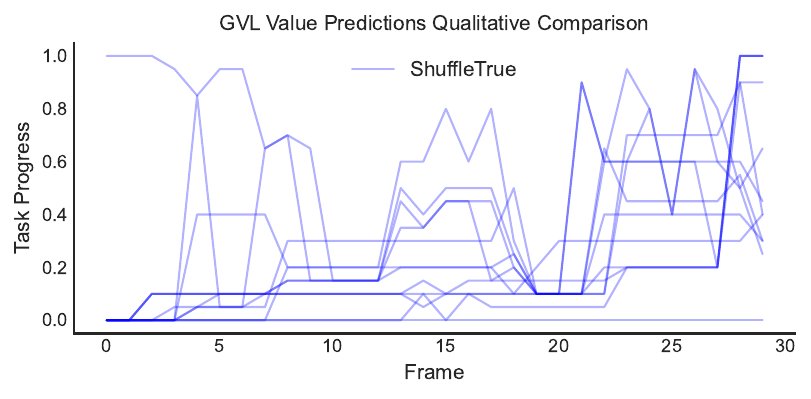}
\includegraphics[width=0.45\textwidth]{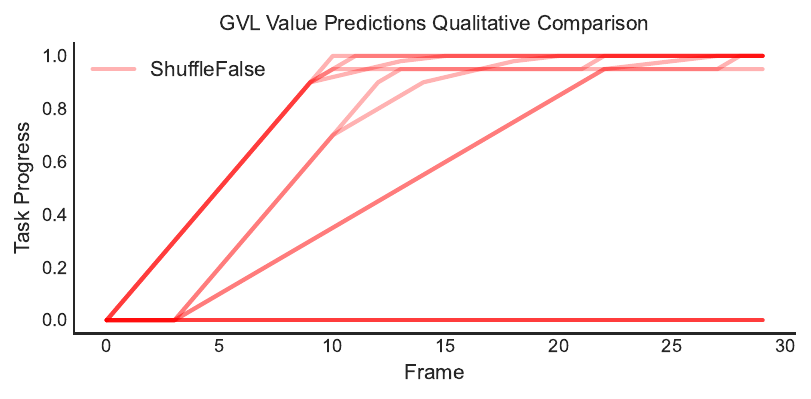}
\caption{\ourmethod without shuffling produces uninformative monotonic values regardless of trajectory quality.}
\label{fig:icv_shuffle_ablations}
\end{figure}

\textbf{\ourmethod and No-Shuffling ablation qualitative comparison.} As shown in~\cref{fig:icv_shuffle_ablations}, \ourmethod generates value predictions that are varied over time; in contrast, without frame shuffling, the predictions all collapses onto a few monotonic patterns. 

\begin{figure*}[t!]
    \centering

    \includegraphics[width=0.68\textwidth]{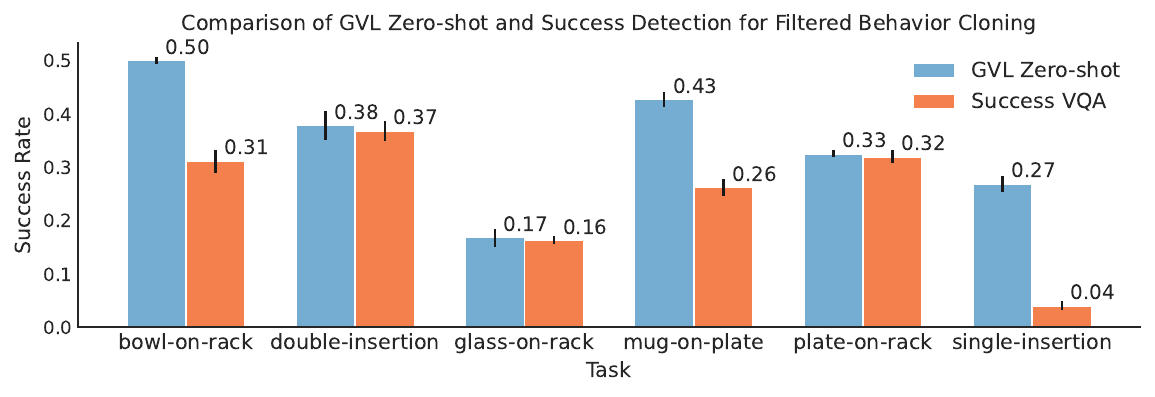}
    \includegraphics[width=0.31\textwidth]{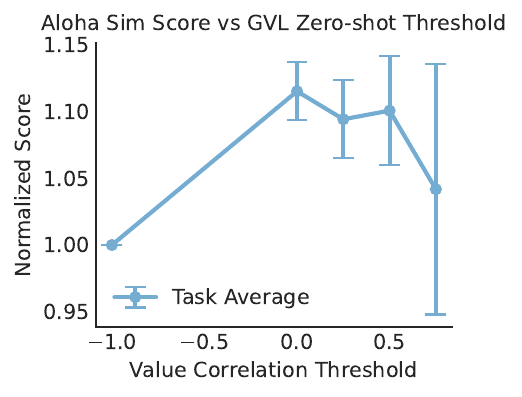}
    \caption{\textbf{Success-Filtered Imitation Learning on ALOHA Tasks.} Left: Using \ourmethod-SD for success-filtered BC substantially outperforms SuccessVQA. Right: \ourmethod-SD is not sensitive to the VOC threshold for improving imitation learning.}
    \label{fig:icv_filtered_bc_zero_shot}
\end{figure*}

\textbf{Zero-Shot Aloha Sim Results.} In \cref{fig:icv_filtered_bc_zero_shot} we include results for \ourmethod-SD zero-shot instead of one-shot. The results are qualitatively similar, where \ourmethod-SD consistently outperforms SuccessVQA, and different VOC threshold values all provide performance gain.

\begin{figure}[h]
\centering 
\includegraphics[width=0.5\textwidth]{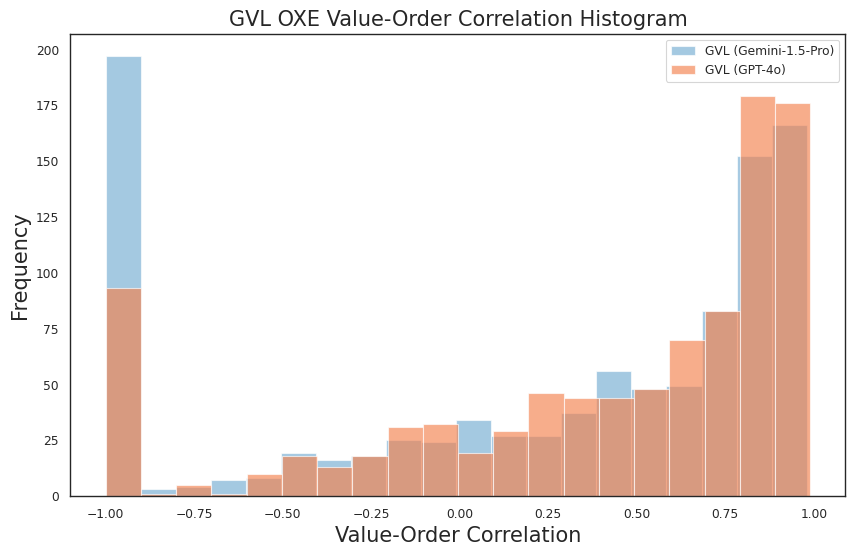}
\caption{\ourmethod has comparable performances with different backbone VLMs; the main difference is in the backbone model's refusal rate and conforming to the response template, which is reflected in the tall bar at $-1.0$.}
\label{fig:vlm_backbone}
\end{figure}

\textbf{Different VLM backbone.} We additionally consider GPT-4o as the backbone VLM to better understand \ourmethod's performance in relation to the backbone VLM model. For evaluation, we plot the histogram of all 1000 (50$\times$20) Value Order Correlation (VOC) scores across all trajectories in Figure~\ref{fig:vlm_backbone}. As shown, \ourmethod, independent of the backbone models, consistently generates VOC scores that heavily skews to the right, indicating that it is able to zero-shot recover the temporal structure hidden in the shuffled demonstration videos, i.e., coherent value predictions. 

\begin{figure}[h]
\centering 
\includegraphics[width=0.49\textwidth]{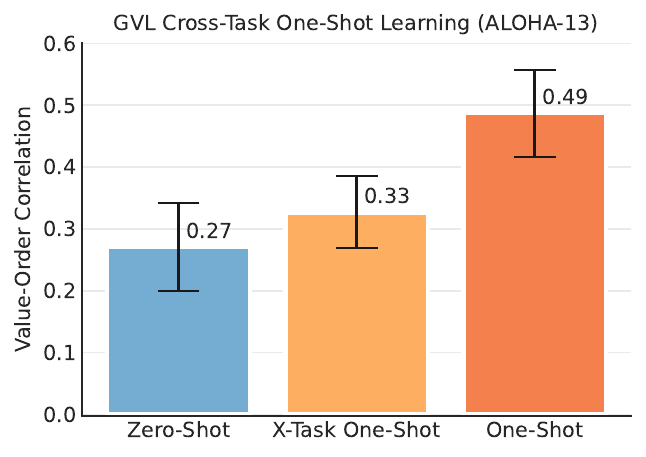}
\caption{\ourmethod demonstrates cross-task in-context learning capability: its value predictions can be improved by value examples from different tasks.}
\label{fig:icv_cross_task}
\end{figure}

\textbf{Cross-task in-context learning.} We investigate whether examples from other tasks can also unlock \ourmethod's in-context learning capability. On the previous ALOHA-13 tasks, we randomly pair up tasks, where we draw one demonstration from one task as the one-shot in-context example for another. Then, we compare VOCs with the original same-task one-shot setup. The results are shown in Figure~\ref{fig:icv_cross_task}. We see that providing examples from a different task is still beneficial, though the improvement is not as much as same-task examples. This is to be expected as intra-task examples still provide clue on the output format as well as a generic notion of task progress, but such information is not specific to the target task. That said, cross-task ICL enables the flexibility of enabling foundation model guidance on a task without any task-specific prior. 

\begin{figure}[h]
\centering 

\includegraphics[width=0.45\linewidth]{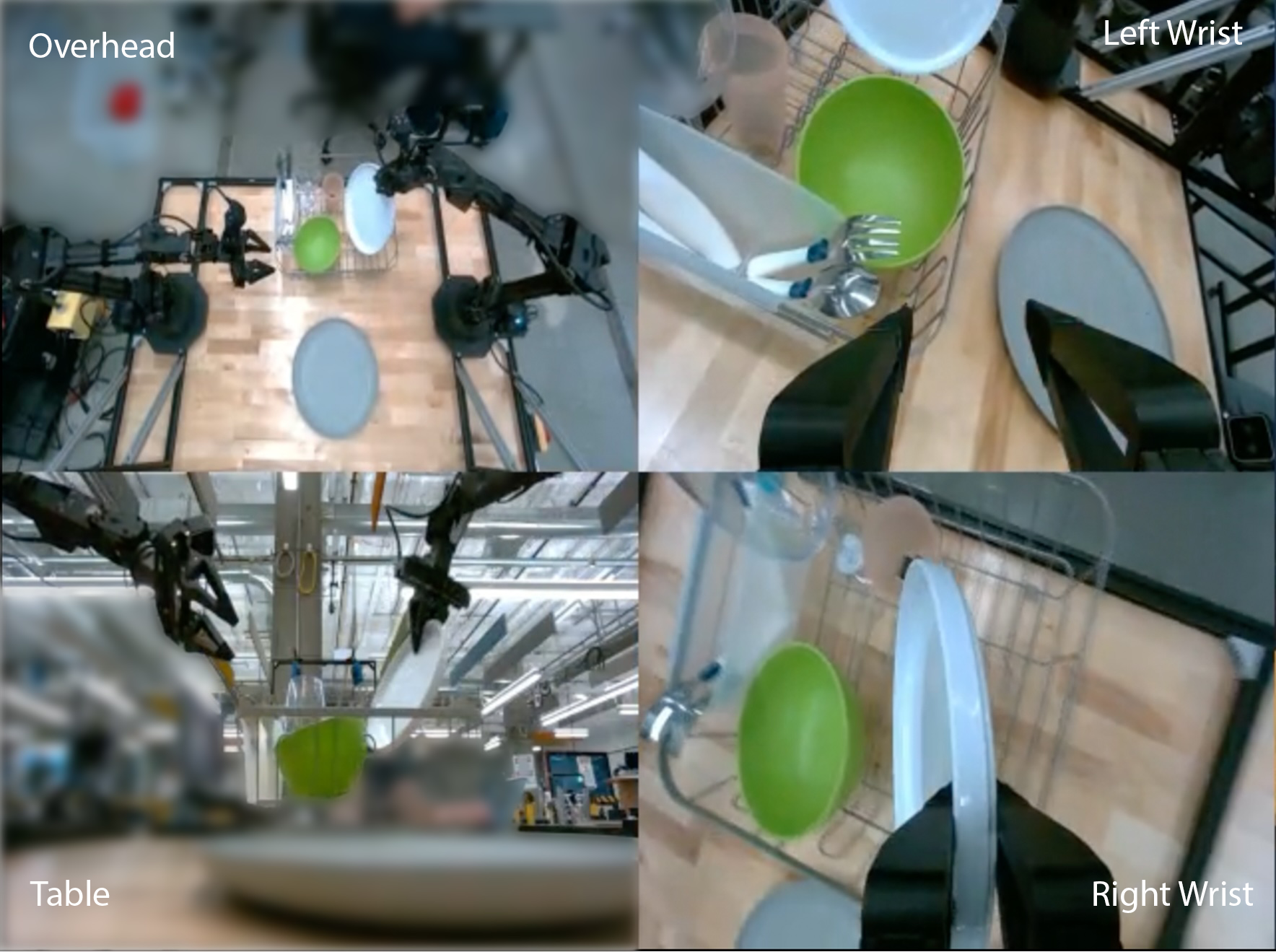}
\includegraphics[width=0.45\textwidth]{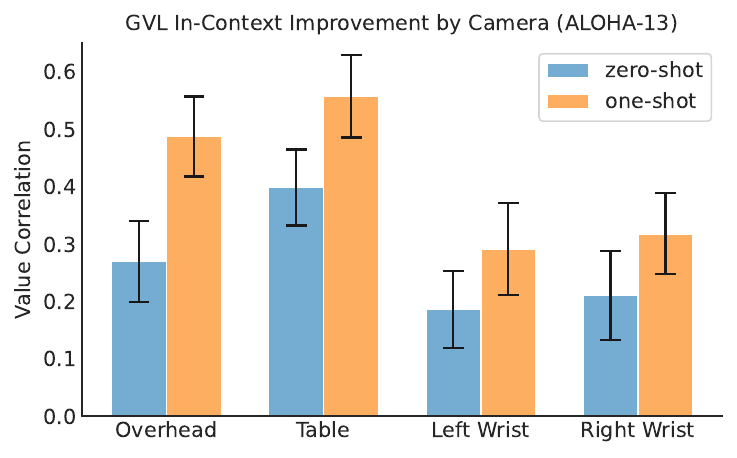}
\caption{\ourmethod works better on more in-distribution table view, but one-shot improvement benefits all camera views.}
\label{fig:icv_camera_ablations}
\end{figure}

\textbf{Does \ourmethod work on different camera viewpoints?} 
On our ALOHA setup, we collected all demonstrations using four camera viewpoints. Besides the top-down view reported in the main experiment above, we test whether \ourmethod remains performant when using alternative viewpoints, especially gripper views that are likely more out-of-distribution with respect to the natural images used for VLM training. The aggregate zero-shot and one-shot results are shown in Figure~\ref{fig:icv_camera_ablations}. As seen, on average, \ourmethod zero-shot works best on the Table viewpoint. This is not surprising, as images taken with the front facing table camera are arguably visually closer to naturally captured images used for VLM training. Yet, with in-context examples, \ourmethod consistently improves on all camera viewpoints. In practice, this means that \ourmethod is robust to camera viewpoints -- even when a camera viewpoint is determined to be sub-optimal post-hoc, practitioners can make up for that by simply providing few in-context examples.

\begin{figure}[h]
\centering 
    \includegraphics[width=\textwidth]{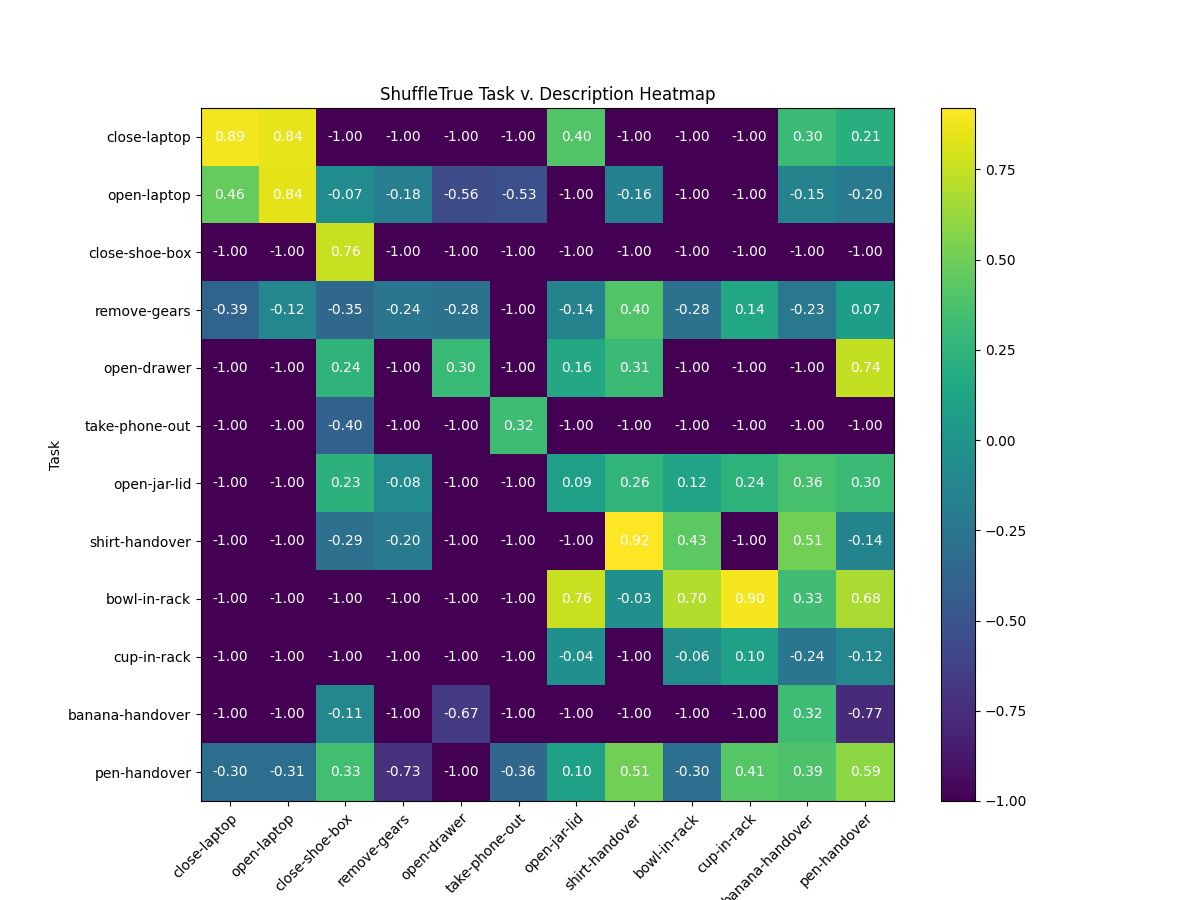}
    \caption{\ourmethod VOC for video and language description pairs. Shuffling enables \ourmethod to pay attention to the language task description in order to faithfully predict observation values.}
    \label{fig:shuffle_true}
\end{figure} 

\begin{figure}
    \includegraphics[width=\textwidth]{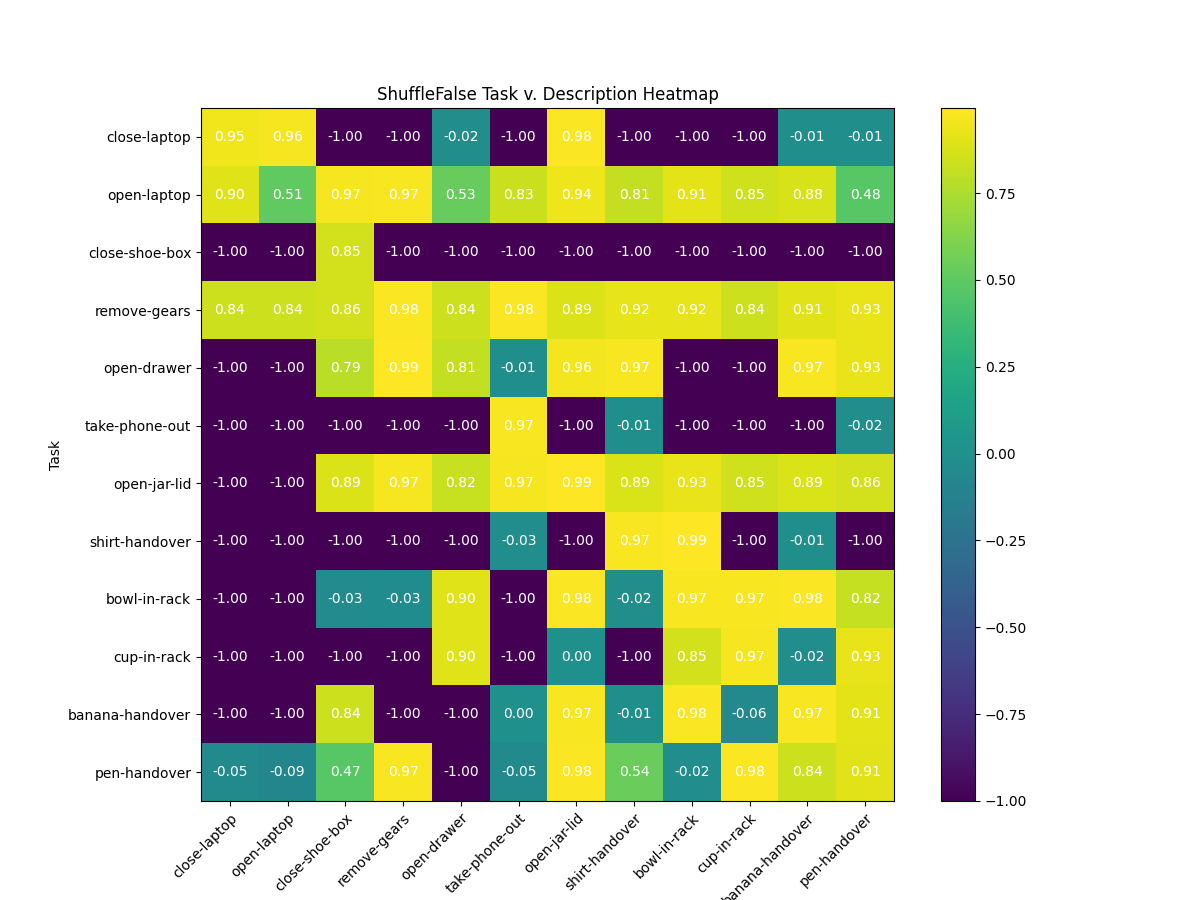}
    \caption{No-shuffling ablation VOC for video and language description pairs. Removing shuffling makes VLM output high VOCs independent of task descriptions.}
    \label{fig:shuffle_false}
\end{figure}

\textbf{Does \ourmethod pay attention to the task specification?}
To validate that \ourmethod is not merely recovering the temporal coherence in the shuffled input video but actively tracking visual progress according to the task language command, we compute the VOC scores for every combination of task input video and language description in the ALOHA-13 split. The heatmap visualization of the average VOC for every pairing is illustrated in~\cref{fig:shuffle_true} and~\cref{fig:shuffle_false} for \ourmethod and the no shuffling ablation. On 9 out of 13 tasks, \ourmethod achieves the highest VOC when the input video and the task description matches; in many unmatched cases, the model simply refuses to output value predictions, stating that the frames and the language description are not related. In contrast, when we do not shuffle the input frames, the quality dramatically drops.

\end{document}